\documentclass{article}

\usepackage[preprint]{neurips_2026}

\usepackage[utf8]{inputenc} % allow utf-8 input
\usepackage[T1]{fontenc}    % use 8-bit T1 fonts
\usepackage{hyperref}       % hyperlinks
\usepackage{url}            % simple URL typesetting
\usepackage{booktabs}       % professional-quality tables
\usepackage{amsfonts}       % blackboard math symbols
\usepackage{nicefrac}       % compact symbols for 1/2, etc.
\usepackage{microtype}      % microtypography
\usepackage{xcolor}         % colors
\usepackage{todonotes}

\usepackage{graphicx}
\usepackage{amsmath}
\usepackage{amssymb}
\usepackage{mathtools}
\usepackage{amsthm}
\usepackage{algorithm}
\usepackage{algorithmic}
\usepackage{caption}

\newcommand{\stdpm}[1]{\text{\,\scriptsize\textpm\,#1}}
\newcommand{\sigimprove}[1]{\textcolor{green!60!black}{\textbf{#1}}} % significant improvement

\title{The FIL Hypothesis:\\ Inductive Biases Help with Kernel Engineering}

\setcitestyle{numbers,square}
\begin{document}

\author{
  Nikolai Rozanov$^{1,3}$\thanks{\textbf{Correspondence to:} Nikolai Rozanov, \texttt{nikolai.rozanov13@imperial.ac.uk}} \and
  Subhabrata Dutta$^{2}$ \and
  % Marek Rei$^{3}$ \and
  Preslav Nakov$^{1}$ \and
  Iryna Gurevych$^{2,1}$
}

\maketitle

\vspace{-2.5em}

% Affiliations (printed directly after the title)
\begin{center}
  \begin{tabular}{c}
    % $^{*}$Equal contribution \\
    $^{1}$NLP Department, MBZUAI, Abu Dhabi, UAE \\
    $^{2}$Ubiquitous Knowledge Processing, TU Darmstadt, Darmstadt, Germany \\
    $^{3}$Department of Computing, Imperial College London, London, UK \\
    % $^{4}$School of ZZZ, Institute of WWW, Location, Country
  \end{tabular}
\end{center}

% \vspace{0.5em}
% \textbf{Correspondence to:} Nikolai Rozanov \texttt{nikolai.rozanov@gmail.com}

% \maketitle

% Abstract: FINAL
\begin{abstract}
% As AI systems are increasingly getting into the real world, they learn from interacting with the environment in a prediction-reward cycle. Traditionally, this loop was very fast. Our main hypothesis, however, is that the future of AI systems will increase this \textit{Feedback Information Loop} (FIL), i.e. the time between an AI system's prediction and the reward signal. 
% Consequently, an increased FIL poses a fundamental barrier to purely data driven machine learning approaches.
The \textit{Bitter Lesson}, which posits that general-purpose methods that scale with computation and data ultimately outperform those with built-in human knowledge, has become a dominant paradigm in the era of Large Language Models. We revisit this principle by observing a new and critical scaling dimension: the duration of the \textit{Feedback Information Loop} (FIL), the time required for a system to receive a verification signal after generating a prediction. 
Most historic successes in Artificial Intelligence (AI) have benefited from near instantaneous feedback (e.g., games or classification tasks), but we argue that future AI applications in science and the physical world will inherently involve FILs ranging from hours to weeks. This trend poses a fundamental scaling limit, as obtaining enough verification steps required by purely data-driven methods becomes practically impossible. 
Additionally, we propose a method that is orthogonal to purely data-driven approaches,
based on human-inspired expert knowledge. The method relies on inductive biases and constraining the solution space. 
We provide an initial validation of the hypothesis and the method, by studying the real-world GPU programming task, a domain with non-trivial FIL, and demonstrate that incorporating inductive biases yields superior performance over data-driven approaches. The code is released under: \url{https://github.com/ai-nikolai/robust_kernelbench}.
% To provide an initial validation of the hypothesis and method, we study two challenging domains, real-world GPU programming tasks and a subset of scientific paper writing, both domains with a non-trivial FIL and demonstrate that incorporating inductive biases yields superior performance over data-driven approaches.
\end{abstract}

\section{Introduction}
% Over the last 7 years, ever since the first publication of 
% The \textit{Bitter Lesson} \cite{sutton2019bitter}, published nearly seven years ago, states that AI systems that benefit from scaling with computation and data are those that outperform other approaches ``by a big margin'', notably those that rely on human inspired inductive biases \cite{battaglia2018relationalinductivebiasesdeep}. 
The \textit{Bitter Lesson}, published nearly seven years ago, states that Artificial Intelligence (AI) systems that benefit from scaling with computation and data are those that outperform other approaches ``by a big margin'', notably those that rely on human-inspired inductive biases \cite{battaglia2018relationalinductivebiasesdeep}. 
Since the publication, this lesson has seen even more empirical evidence with Large Language Models (LLMs) not only when solving various classical Natural Language Processing (NLP) tasks \cite{wang2019gluemultitaskbenchmarkanalysis, wang2020supergluestickierbenchmarkgeneralpurpose} that seemed extremely hard only a decade ago, but also when solving tasks such as problems from the Math Olympiad \cite{Hubert2025AlphaProof}, the International Collegiate Programming Competition (ICPC) \cite{zou2025liveoibench}, as well as \emph{Agentic AI} tasks such as Web Navigation, Robotics and many others \cite{rozanov2025stateactenhancingllmbase, yao2023reactsynergizingreasoningacting}. A key method behind these successes was the reinforcement learning-based scaling \cite{Guo2025DeepSeek} of LLMs. 

In line with the \textit{Bitter Lesson} these approaches benefit from scaling with large amounts of data and computational effort. 
% A key ingredient for these approaches is the ability to actually receive a signal, e.g. a reward signal in Reinforcement Learning. 
As can be seen in Table~\ref{tab:data_needs}, previous state-of-the-art approaches scaled and exceeded human performance with data scales ranging from around $O(10^6)$ simulations or $O(10^{12})$ tokens of training data.
Contrary to existing views, we propose a new dimension to think about such systems: the \textit{Feedback Information Loop} (FIL). We define the Feedback Information Loop as the duration for an AI system to receive `feedback' after it has made a prediction. This can be the reward (+1, 0 or -1) when playing chess or Go, the accuracy in classification tasks, a numerical reward on LLM generated responses and so on.
The crucial observation is that most, if not all, existing problems enjoyed a very short Feedback Information Loop; given an AI response and a desired gold answer it is very quick to compute whether the AI system is right. For example, in chess it is very quick to compute if the position is a draw, a win, or a loss; likewise, in natural language processing, one can quickly tell whether the answer is correct, etc. 

Looking into state-of-the-art and future AI application areas, it is evident that the Feedback Information Loop is only bound to increase.
% which will only further hinder the purely data-driven approaches.
This can be easily seen in applications such as the AI Scientist \cite{lu2024aiscientistfullyautomated} (e.g.,~designing automatic GPU code, or machine learning experiments) or AI for the physical world (e.g.,~designing a chemical synthesis, or a new aerodynamic design), which will take several hours to several weeks to verify, as shown in Table~\ref{tab:data_needs_2}. The scaling to the required\footnote{The number of required data points for these methods to scale is consistently very large. $10^6$ is taken as a lower estimate given the existing problems.} $O(10^6)$ data points in such cases would yield practically impossible timelines for purely data driven approaches. This can be clearly seen in Table~\ref{tab:FIL_calculation}, e.g.,~if a single verification takes an hour to verify, then $10^6$ batches would take around 114 years of verification, if it takes a day that would be nearly 3,000 years of verification, while if the verification takes a week it would take almost 20,000 years to verify. Hence purely data-driven approaches are becoming prohibitively expensive in the extended FIL setting.

Previous approaches did not openly identify this issue; however, they can be seen as implicit attempts to solve the long FIL problem.
% to tackle this problem\footnote{Previous worked observed and tackled reduced variants of the FIL hypothesis, often attributing the delayed feedback to a specific problem.} 
Specifically, some existing work has attempted to overcome the orthogonal but related problem of sparse feedback signal  \cite{NEURIPS2022_266c0f19_Sparse_rewards, burda2018explorationrandomnetworkdistillation}; however, the underlying problems have short FIL and do not fundamentally overcome the problem. The most direct comparison is existing work that aims to simulate the underlying process and thereby to implicitly approximate the Feedback Information Loop, such as World Models, model-based Reinforcement Learning, reward modeling, and other forms of modeling the environment or the feedback \cite{NEURIPS2018_2de5d166_world_models_2018, Hafner2025DreamerV3}. Such approaches are limited, however, as shown in the latest comprehensive work \cite{10.1145/3746449_world_model_survey}, specifically due to (1)~the need for training data, and (2)~the so-called ``simulation to real'' gap, which represents the aleatoric and epistemic uncertainty of the models. When looking at long FIL, we see that these two issues with environment models would only amplify. Hence, data-driven approximation techniques also do not represent a fundamental solution.

Our \textbf{main hypothesis}, the FIL hypothesis, is that ``as the Feedback Information Loop increases in duration, it will pose a fundamental limit to purely data-driven methods'', thereby extending the \textit{Bitter Lesson}. 

% are not sufficient to solve the problem. 
% TODO: add why existing approaches are not sufficient.

% PROPOSAL Version from 31.03.2026
% Our \textbf{main proposal}, is that: ``human
% % \footnote{Human inspired, in this context, means expert knowledge, as well as other social or physiological inspiration} 
% (or nature) inspired 
% inductive biases\footnote{We define inductive biases as constraints on the solution search space. These can be inspired by human intuition or other (natural) systems.}
% % and other forms of co-creation \cite{HAICO_SUBHABRATA_10.1162/COLI.a.19} 
% and machine learning systems should pose a renewed avenue of fundamental and mainstream machine learning research''.

% PROPOSAL new version
Inspired by this observation and hypothesis, we propose a method based on inductive biases. We define inductive biases as constraints on the solution search space. These are often inspired by a human expert , which is orthogonal to purely data-driven approaches.
%TODO: Need to think about the phrasing of our hypothesis...Need to update this paragraph...
In order to show an initial validation of the FIL-Hypothesis and our proposal to use human expert-based inductive biases,
% \footnote{We define inductive biases as constraints on the solution search space. These can be inspired by human intuition or other (natural) systems.}
 we chose a real-world problem that exhibits the slow Feedback Information Loop trend and designed an algorithm with human-inspired solution constraints. Specifically, we chose a challenging GPU programming task \cite{ouyang2025kernelbenchllmswriteefficient} in line with the AI Scientist paradigm, where AI can help the human solve real and challenging scientific tasks. 
Our results demonstrate that even use cases with only several minutes in their Feedback Information Loop duration already show signs of the improvement via human-inspired algorithm design; we outperform the state-of-the-art baseline methods by introducing our inductive bias across different model sizes, settings and evaluation measures.

\begin{table*}[htbp]
\centering
\caption{Training data scale of the selected machine learning systems.}
\label{tab:data_needs}
\begin{tabular}{p{2cm} p{2cm} p{2cm} p{2cm} p{2cm}}
% \begin{tabular}{c c c c c}
\toprule
\textbf{System} & \textbf{Primary Domain} & \textbf{Type of Training Data} & \textbf{Quantity} & \textbf{Feedback Information Loop (FIL)} \\
\midrule
AlphaGo Zero \cite{Silver2017_AlphaGoZero} & Specialised RL Algorithm & GO Games &  $\sim$4.9M $O(10^6)$ & $<1$ second\\
\hline
DeepSeek-R1 \cite{Guo2025DeepSeek} & Large Language Reasoning Model & Verifiable Problems & $\sim$5M $O(10^6)$ & $<1$ second \\
\hline
Dreamer V3 \cite{Hafner2025DreamerV3} & General RL Agent & Game Steps & $\sim$100M $O(10^8)$ & $<1$ second \\
\hline
AlphaProof \cite{Hubert2025AlphaProof} & Automatic Mathematical Prover & Formal Problems & $\sim$80M $O(10^7)$ & $<1$ seconds \cite{santos2025kiminaleanserverhighperformance} \\
% \hline
\bottomrule
\end{tabular}
\end{table*}

% Therefore we also propose that methods that use inductive biases should receive a renewed attention in the research community.
% \todo[inline]{TODO}

% IMPORTANT: TODO: I guess we should also speak about that our "Inductive Bias" method is also a contribution, because it is based on the insight from FIL. Ie. that you need to shorten FIL to succeed with "Data driven approaches". Hence we decompose the loss to reduce the FIl cycles. 

% TODO: update table with:
% 1. Concrete data amounts (incl. batchsize)
% 2. Verification criteria
% 3. Verification Duration
% 4. Citations of the actual papers...

\section{The Feedback Information Loop}
\label{sec:background}

The core scaling dimensions in the last years of machine learning research have been (1)~data and (2)~compute. In our work, we introduce a third critical one: the Feedback Information Loop.

\subsection{Defining the Feedback Information Loop}
\label{subsec:FIL_motivation}
\begin{figure}[htbp]
    % --- Left Minipage for the Table ---
    
    \begin{minipage}[c]{0.48\textwidth}
        \centering
        % NeurIPS rule: Table caption BEFORE the table
        \captionof{table}{The calculation of the \textbf{total verification time} given different Feedback Information Loop (FIL) times for $10^6$ and $10^3$ data points, respectively. $\mathcal{O}(10^6)$ represents the case where traditionally that was\ sufficient data for super-human performance. $\mathcal{O}(10^3)$ represents an ``idealistic'' data-efficient scenario.}
        \label{tab:FIL_calculation}
        \begin{tabular}{l|cc}
            \toprule
            \centering\textbf{FIL} & \textbf{$\mathcal{O}(10^6)$} & \textbf{$\mathcal{O}(10^3)$} \\
            \midrule
            1 millisecond & 17 minutes & 1 second \\
            1 second & 12 days & 17 minutes \\
            1 minute & 694 days & 17 hours \\
            1 hour & 114 years & 42 days \\
            1 day & 2,738 years & 2.7 years \\
            1 week & 19,165 years & 19 years \\
            1 month & 83,136 years & 82 years \\
            \bottomrule
        \end{tabular}
        % Note: No \caption here, as it appears above.
    \end{minipage}
    \hfill % Adds flexible space between the minipages
    % --- Right Minipage for the Figure ---
\begin{minipage}[c]{0.48\textwidth}
        \centering
        % Ensure width is relative to minipage, not full textwidth
        \includegraphics[width=0.65\linewidth]{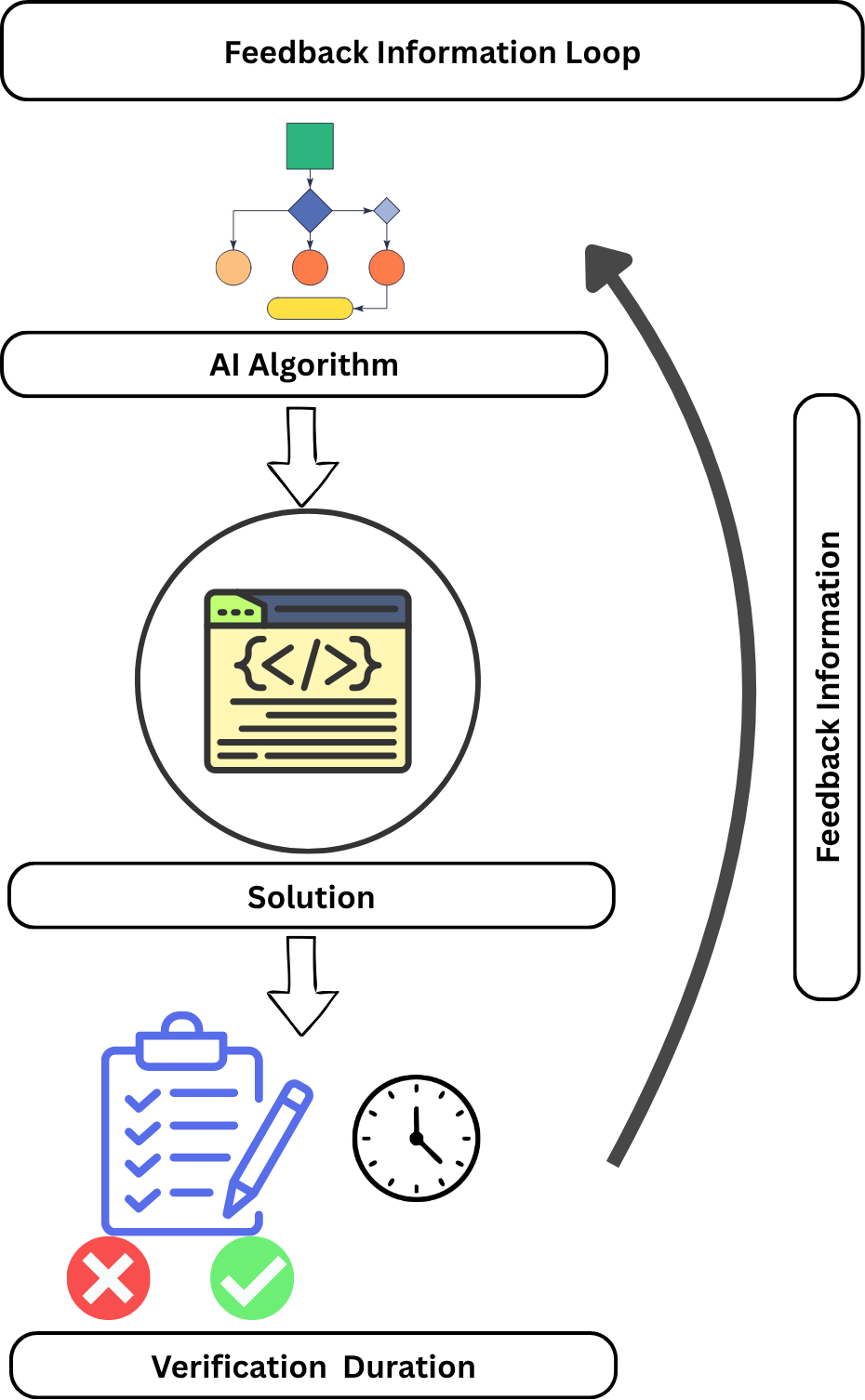}
        % NeurIPS rule: Figure caption AFTER the graphic
        \captionof{figure}{Illustration of the different stages of the Feedback Information Loop (FIL). Step 1:~The AI algorithm produces a solution. Step 2:~The solution is verified. Step 3: The verification signal is used by the algorithm.}
        \label{fig:FIL}
    \end{minipage}
\end{figure}

We define the Feedback Information Loop as the time it takes for an AI system to receive feedback on its prediction or a batch of predictions, as seen in Figure \ref{fig:FIL}. There are three main stages in the Feedback Information Loop: (1)~the actual prediction mechanism, (2)~the verification of the prediction, (3)~the (potential) use of the feedback information. Stages 1 and 3 are dependent on the algorithmic choices. In the case of LLMs, for example, the actual prediction happens via the forward pass of the LLM, while the usage of feedback can happen either via updating the prompt or via updating the parameters. 

% \begin{figure}[ht]
%   \vskip 0.2in
%   \begin{center}
%     \centerline{\includegraphics[width=0.4\linewidth]{images/FIL3.png}}
%     \caption{
%       Feedback Information Loop
%     }
%     \label{fig:FIL}
%   \end{center}
% \end{figure}

% The entire block (figure and table) floats together as a single 'figure' object.

The total FIL duration is calculated by adding the time for each stage. Since stages 1 and 3 depend on the algorithm; algorithmic improvements, hardware improvements and other techniques can be used to reduce this time.\footnote{Naturally, for stages 1 and 3, it is sometimes not trivial to continue reducing their duration and perhaps other fundamental limits might exists, only further amplifying the FIL.} Therefore, for the purpose of our paper, we simplify the FIL duration to the duration of stage 2 only, which mainly depends on the underlying problem to be solved and is fully independent of the choice of algorithm.

Existing AI problems that enjoyed widespread recognition and success, such as protein folding prediction \cite{Jumper2021_alphafold}, the game of GO \cite{Silver2017_AlphaGoZero}, Large Language Models for mathematical and computational competitions \cite{Guo2025DeepSeek, Hubert2025AlphaProof}, and many more have a Feedback Information Loop (FIL) of less than a second. Therefore, training for $O(10^6)$ iterations takes around a day (in terms of verification costs). However, for emerging and future AI applications, as shown in Table \ref{tab:data_needs_2} and Appendix Table \ref{tab:data_needs_2_detailed} training for the same number iterations would take unrealistic time: for example one hour verification per batch leads to 114 years of total verification time.

% CURRENTLY WRITING:
\subsubsection{Data Needs and Scaling Laws}
The scaling laws that underpin modern large language models implicitly assume the existence of vast, static datasets \citep{chinchilla_10_5555_3600270_3602446}. The duration of the Feedback Information Loop (FIL) introduces a critical new dimension to these scaling considerations. The well-established \emph{Chinchilla} scaling laws posit that for compute-optimal training, model size and training tokens should be scaled equally: doubling model size requires doubling the number of tokens \citep{chinchilla_10_5555_3600270_3602446}. 

This problem is exacerbated for domains like code generation or physical process modeling, which emerging evidence suggests reside in a ``more data-hungry regime'' than natural language \citep{luo2025scalinglawscodedatahungry}. In reinforcement learning, this has often been approached with replay buffers \cite{NEURIPS2019_5c48ff18_search_on_replay_buffer}, which allows the reuse of previous data, but still requires new data to be collected. Recent work on data-efficient reasoning has demonstrated that a model can achieve state-of-the-art performance on competition-level mathematics with as few as 817 curated training examples \citep{ye2025limo}. Crucially, however, achieving these results still requires 15 training epochs over this small dataset, meaning that the model sees each example 15 times and performs over 12,000 total training steps \citep{ye2025limo}. While this represents a dramatic improvement in sample efficiency, it does not eliminate the underlying requirement for multiple verification signals per data point. For a real-world scientific task with a FIL of days or weeks, the cumulative time required to complete this number of  training iterations
% , let alone the iterative refinement process needed to achieve robust generalization 
remains prohibitive. Even if we optimistically assume that a scientific task requires only the $O(10^4)$ verification steps necessary for a purely data-driven approach, the cumulative waiting time 
% accumulated from repeated feedback loops over weeks or months per iteration 
renders this strategy infeasible, as at 1 hour verification time per batch, this would take a year, and for a task that takes 1 day, this would take 24 years of verification, and for a task that takes 1 week, this would take over a century of verification time.

Thus, while data-efficient methods narrow the gap, they do not overcome the fundamental barrier imposed by a long feedback loop. Consequently, the extended feedback delays inherent in scientific and physical applications demand a paradigm shift away from purely data-driven approaches.

% Want to include:
% 1. Chinchilla (i.e. bigger model -> more data)
% 2. Scaling plots / table (different data amounts, vs. different FIL)
% 3. That some problems might need more data than text (e.g. coding; could also be true for physical process)
% 4. Even at the 1000 datapoints lower-bound would yield to long waiting times and it's not very realistic... 

% \todo[inline]{TODO}
% We want to describe the feedback information loop in a bit more detail maybe add a graphic?

% \begin{table*}[htbp]
\begin{table}[htbp]
\centering
\caption{Selected Tasks with Prolonged Feedback Information Loop.}
\label{tab:data_needs_2}
% \begin{tabular}{p{3.8cm} p{3.8cm} p{3.2cm}}
% \begin{tabular}{p{3cm} p{3.7cm}}
\begin{tabular}{l|l}
\toprule
\textbf{Domain} &  \textbf{Feedback Information Loop (FIL)} \\
\midrule
GPU Kernel Programming \cite{ouyang2025kernelbenchllmswriteefficient}  &
% Code Compilation and Performance Evaluation & 
$\sim$2 - 30 minutes\\
\hline
Machine Learning Engineering \cite{chan2025mlebench}   & 
% Datascience and Machine Learning Challenges  & 
$\sim$2 minutes - 1 hour \\
\hline
OS Kernel Engineering  & 
% Compiling new OS Kernels and checking if a specific metric has improved (performance, memory, security) & 
$\sim$ 5 minutes \cite{borges2025linuxkernelconfigurationsscale} \\
\hline
Real Chemical Synthesis   & 
% Checking if a proposed synthesis plan yields a given substance & 
$\sim <$ 1 hour - several weeks \cite{Volk2023_alpha_flow_chemical_synthesis} \\
\hline
Automatic Large Language Model Research &
% Checking whether a given LLM recipe (training data and training pipeline) yields an improved LLM & 
$\sim <$ 1 day - several months \cite{Pretraining_runs_100k_100days} \\
% \hline
\bottomrule
\end{tabular}
\end{table}
% \end{table*}

% \input{sections/2_background}

\section{Method Based on Inductive Biases}
Our method is motivated by the slow Feedback Information Loop. Previous approaches have attempted to overcome this challenge by modeling the environment and simulating it, or finding ways to reuse data. Our proposed direction is orthogonal. We propose to imbue deep domain knowledge into the system where the FIL duration is large. 
% In this study we propose a method based on this notion. 
Specifically, our method focuses on test-time search. We propose to break up the test-time search into subproblems based on how human experts solve the task. Interestingly, this also leads to reduced FIL durations for each individual stage of the problem.

% \subsection{Feedback Information Loop Analysis}
% \label{sec:fil_analysis}
% \todo[inline]{TODO: Feedback information loop, here we want to communicate and derive the section more about how long each task takes etc...}
% The Feedback Information Loop (FIL) is important to understand better.
% ...

% \subsection{Inductive Biases}

% \subsection{Our Main Hypothesis and Proposal}
% \todo[inline]{TODO}

% As previously discussed and shown in Table \ref{tab:data_needs_2}, we hypothesize that purely data-driven approaches will find limitations, especially as FIL duration grows. Therefore our main hypothesis is that in such cases human intuition and knowledge will become highly important in guiding the machine learning system in finding solutions. 

In line with the idea of AI searching for solutions in a (hypothesis) space \cite{HAICO_SUBHABRATA_10.1162/COLI.a.19}, we define inductive biases as a constraint of the search space for the AI system, similarly to previous work such as \cite{Mitchell2007TheNF, battaglia2018relationalinductivebiasesdeep, sahnan2026cofactcheckerframeworkhumanaicollaborative}. While the exact notion of inductive biases might differ between previous work, the commonality is the ``biasing'' or constraining an AI system to allow for better generalization.

% In this section we put forward that perhaps research into inductive biases should be conducted as a viable method of finding solutions to problems that have high FIL. Specifically, we want to argue that this is a large open field... Ie. instead of research more and more scale with data, it's about researching how to include expert knowledge and inductive biases better into the AI models and systems.

\subsection{Inductive Bias Search}
% \todo[inline]{TODO}

% Starting with our observation that FIL duration has a big impact on data availability and also limitations of feedback to the AI system.
% % \todo{incomplete sentence?} 
% We propose a method that both incorporates human knowledge as well as reduces the FIL duration, in each individual stage.

% Specifically, given a problem like GPU programming, machine learning research, chemical synthesis, etc.; we 
We start with the observation that humans typically break up a complex process into smaller stages. For example, in the case of GPU programming, when writing a specific GPU kernel, we first try and make it run, then check for correctness and next, we look for ways to  improve the code, and we iterate this sequence; or, in chemical synthesis, we synthesize prerequisite chemicals first before combining them in a final synthesis stage.

Following this principle, we propose \textbf{Inductive Bias Search}. Similarly to previous test-time scaling approaches \cite{NEURIPS2023_271db992_tree_of_though, rozanov2025stateactenhancingllmbase}, our method requires iterative search in the hypothesis space. Additionally, we use verifiable rewards similarly to Reinforcement Learning with Verifiable Rewards (RLVR) \cite{deepseekai2025deepseekr1incentivizingreasoningcapability} to determine the different stages. The core of our method lies in the decomposition of these rewards and stages. Existing approaches group complex rewards into a single scalar value \cite{baronio2026kevin}, often via some weighting. In our case, we introduce an ordered reward function that consists of limiting the solution space at every stage. Schematically the difference compared to classical (existing) approaches and our method can be seen in Appendix \ref{app:multi_stage_search}, Figure \ref{fig:classic_search} for the classical approach and Figure \ref{fig:multi_stage_search} for our method.

% %%%%%%%%%%%%%%%%%%%%%%%%%%%%%%%%%%%%%
% 
% 
% MAIN INPUT TO THE METHOD SECTION

% New...
\subsubsection{Formalizing the Inductive Bias Search under Feedback Constraints}
% \label{sec:formal_multi_stage}
\label{sec:formal_def_multi_stage}

Let a machine learning task be defined by an input space $\mathcal{X}$, a solution space $\mathcal{Y}$, a reference solution $y_{\text{ref}} \in \mathcal{Y}$, and a final verification metric $\mathcal{V}(y, y_{\text{ref}}): \mathcal{Y} \times \mathcal{Y} \rightarrow \mathbb{R}$ (which we denote simply as $V(y)$ when the reference is fixed). The goal is to find
\[
y^* = \arg\max_{y \in \mathcal{Y}} V(y).
\]

\paragraph{Ordered decomposition.}
An \textbf{ordered decomposition} of $V$ is a non-increasing sequence of sub-metrics $\{V_i\}_{i=1}^N$ such that $V_i(y) \ge V_j(y)$ whenever $i < j$ and $V_N = V$. Intuitively, each $V_i$ represents an ``easier'' sub-objective that is less stringent than the final metric. For each stage $i$, we associate a performance threshold $t_i$ and a (soft) inductive bias $\mathcal{B}_i$ that guides the search toward solutions satisfying $V_i \ge t_i$.

\paragraph{Feedback budget constraint.}
Let $\tau$ denote the duration of a single Feedback Information Loop (FIL) the time from generating a candidate solution to receiving its verification signal. Given a total time budget $B$ and the ability to run $k$ independent verification processes in parallel, the maximum number of verifiable evaluations is
\[
M = \frac{k \cdot B}{\tau}.
\]
If $p$ is the probability that a randomly sampled candidate $y \sim \mathcal{Y}$ (under the model’s base distribution) meets the final performance threshold, then under independent sampling the probability of success after $M$ trials is
\[
P_{\text{success}} = 1 - (1-p)^M.
\]
For large $M$, even a small $p$ yields high success probability. However, when $M$ is severely limited (as in long-FIL settings), the mitigation is to increase $p$ itself i.e., to bias the sampling distribution toward high-quality solutions without requiring many verification steps.

\paragraph{Inductive conditioning}
Inductive bias search achieves this by decomposing the final objective into intermediate constraints. Let $\{V_i\}_{i=1}^N$ be \textbf{ordered decomposition} corresponding to progressively stricter constraints, where $V_N = V$ denotes the final objective. For simplicity, we assume each $V_i$ defines a binary predicate indicating whether a candidate satisfies a given stage.

Each evaluation function $V_i$ induces a feasible subset of the hypothesis space:
\[\mathcal{Y}_i = \{ y \in \mathcal{Y} \mid V_j(y) = 1 \ \forall j \le i \}\]

By construction,
\[\mathcal{Y} = \mathcal{Y}_0 \supseteq \mathcal{Y}_1 \supseteq \mathcal{Y}_2 \supseteq \dots \supseteq \mathcal{Y}_N
\]

\paragraph{Increasing the probability of success} 
Inductive bias search is designed to improve the success probability, by constraining solutions at stage i to the respective feasible hypothesis set $\mathcal{Y}_i$. We consider the unconditional success probability for stage i:
\[
p_i = \mathbb{P}(V_i(y) = 1 \mid y \in \mathcal{Y})
\]
and the conditional success probability for stage i, corresponding to our inductive bias search:
\[
\tilde{p}_i = \mathbb{P}(V_i(y) = 1 \mid y \in \mathcal{Y}_j, j<i)
\]

Therefore, we have $p_i \le \tilde{p}_i, \forall i$, since $\mathcal{Y}  \supseteq \mathcal{Y}_i$. 
Therefore, we expect that successive searches yield a higher final stage probability:

\begin{align}
   &\tilde{p}_0 \le \tilde{p}_1 \le \dots \le \tilde{p}_N \le 1
\end{align}

A step by step of the method is given in Appendix \ref{app:multi_stage_search}, Algorithm \ref{alg:multi_stage_search}.

\subsubsection{Construction of the Ordered Decomposition}
The critical stage of the algorithm \ref{alg:multi_stage_search} is the human expert contribution of the ordered decomposition $\left \{ \mathcal{V}_i \right \}_{i=1}^N$, thresholds $\mathcal{T}$ and inductive biases $\mathcal{B}$. Importantly, this stage requires the creation of additional sub-metrics $\mathcal{V}_i$, which often requires creating human-crafted verifiers for these stages. For example in a coding task this could be measurement of specific compilation errors. Secondly, this requires the setting of appropriate thresholds for the sub-metrics as to what counts as a successful passing of the given metric. To simplify the threshold setting we use only binary sub-metrics in our proposed algorithms. Finally, the creation of the (soft) inductive biases depends on the underlying machine learning system; in our study the primary focus is on LLMs and hence the inductive biases come in form of extended prompts and in-context learning.

The choice of the ordered decomposition depends on the task at hand and can be based on the existing process of the human. For example in coding task the debugging pipeline of first fixing syntax errors, then runtime errors, etc.

%TODO: continue on that...

% 
% 
% 
% %%%%%%%%%%%%%%%%%%%%%%%%%%%%%%%%%%%%%

% \begin{figure}[ht]
%   \vskip 0.2in
%   \begin{center}
%     \centerline{\includegraphics[width=\columnwidth]{images/classic_search.png}}
%     \caption{
%       Classical Solution Search
%     }
%     \label{fig:classic_search}
%   \end{center}
% \end{figure}

% \begin{figure}[ht]
%   \vskip 0.2in
%   \begin{center}
%     \centerline{\includegraphics[width=\columnwidth]{images/multi_stage_search.png}}
%     \caption{
%       Inductive Bias inspired Multi-Stage Solution Search (Ours)
%     }
%     \label{fig:multi_stage_search}
%   \end{center}
% \end{figure}

% MAIN FIGURE USED
% \begin{figure*}[ht]
%   \vskip 0.2in
%   \begin{center}
%  % \centerline{\includegraphics[width=0.5\linewidth]{images/classical_search.png}}
%  \centerline{\includegraphics[width=0.7\linewidth]{images/classical_search_v2.png}}
%     \caption{
%       Classical Solution Search
%     }
%     \label{fig:classic_search}
%   \end{center}
% \end{figure*}

% \begin{figure*}[ht]
%   \vskip 0.2in
%   \begin{center}
%     %width=0.9
%     \centerline{\includegraphics[width=0.7\linewidth]{images/multi_stage_search.png}}
%     \caption{
%       Multi-Stage Solution Search (Ours)
%     }
%     \label{fig:multi_stage_search}
%   \end{center}
% \end{figure*}
\section{Experimental Setup}

Overall, our aim with the empirical section is to demonstrate that: 1) slow FIL tasks exist in the wild; 2) demonstrate that inductive-bias based methods, like our inductive bias search can outperform the purely data-drive approaches.

A very practical problem to this end is GPU kernel engineering. This is a very applied and important tasks for AI scientists and engineers. On one hand because it is part of active research itself, such as for ML systems researchers \cite{vllm_10.1145/3600006.3613165}; and on the other hand it is a critical component for AI research itself, as often ML experiments would be infeasible without efficient GPU code. For example libraries such as vLLM \cite{vllm_10.1145/3600006.3613165} or Unsloth \cite{unsloth} facilitate resource constraint training and inference.

% IDEA FOR THIS SECTION
% In this section we want to describe how we want to validate our idea based on a simple, yet hopefully powerful solution...

% We want to motivate the challenge, explain how it fits into FIL and how it is realistic and also perhaps how it can only get harder in the future. 

% (Somehow we should try and say that our aim of the paper is not to necessarily show how to create the best inductive bias method etc. rather just to show the method and demonstrate it on a single task...)

\subsection{The KernelBench Challenge as a Slow-FIL Testbed}
\label{subsec:kernelbench}
% \todo[inline]{TODO}

To ground our investigation in a concrete slow-FIL problem, we adopt \textbf{KernelBench}, a benchmark for evaluating LLMs' ability to write efficient GPU kernels \citep{ouyang2025kernelbenchllmswriteefficient}. KernelBench level 1 comprises 100 real-world PyTorch machine learning workloads, each requiring the generation of a custom GPU kernel that is both functionally correct and provides measurable speedup over the PyTorch baseline \citep{ouyang2025kernelbenchllmswriteefficient}.

The task exhibits several properties that make it an ideal representative of the emerging slow-FIL regime. First, verification is non-trivial: each generated kernel must be compiled, executed, and profiled to assess both correctness and runtime performance. Second, feedback is sparse and delayed: obtaining a reward signal (speedup relative to baseline) requires minutes of compilation and execution time per candidate. Third, the performance landscape is rugged: small semantic differences in kernel code can produce order-of-magnitude differences in runtime, and the mapping from code to performance is not easily approximated.

Critically, the current state-of-the-art on KernelBench is an \textbf{iterative refinement paradigm} that explicitly relies on rapid feedback loops. Frontier reasoning models generate candidate kernels, receive execution and profiling feedback, and refine their solutions over multiple rounds \citep{Baronio2025KevinMR}. This approach is a direct instantiation of the short-FIL assumption.
% : it achieves improved performance precisely because each verification cycle takes only seconds to minutes. 
Additionally, even with this iterative feedback, the best models match PyTorch baseline performance in less than 20\% of cases, and furthermore are failing on simpler metrics like compilation and runtime as well \citep{ouyang2025kernelbenchllmswriteefficient}
% performance degrades sharply as the required speedup threshold increases .

KernelBench therefore provides a controlled setting where the Feedback Information Loop duration is on the order of minutes already sufficient to reveal the limitations of purely data-driven, iterative refinement approaches. 
% As we scale to verification cycles of hours (AI Scientist conducting full ML experiments \citep{lu2024aiscientistfullyautomated}) or weeks (physical synthesis and testing), the inadequacy of the short-FIL paradigm becomes absolute. 
% 
% KernelBench serves as a canary in the coal mine, offering empirical evidence that even modest FIL 
% 
KernelBench offers empirical evidence that even modest FIL durations impose meaningful constraints on existing methods, and a testbed for evaluating inductive biases designed to operate under such constraints.

\subsection{Inductive Bias Test Time Scaling in KernelBench}

In this part we describe the concrete implementation of our proposed method applied to the challenge of GPU kernel engineering.

Following Algorithm \ref{alg:multi_stage_search}, we need to identify the main metric of KernelBench as well as define an ordered decomposition of this metric. 

Given the reference pytorch solution $y$ and the LLM proposed kernel $\hat{y}$, the objective $\mathcal{V}$ for KernelBench is the $fast_k$ metric \cite{ouyang2025kernelbenchllmswriteefficient}, which measures how many solutions are correct and have a speed-up of at least k.

\begin{equation}
    fast_k = 
    % \sum_{i=1}^{100} 
    \mathbb{I}_{\hat{y} \, runs \, corectly}\{ time(y) / time(\hat{y}) >k \}
\end{equation}

\subsubsection{Ordered Decomposition of Metrics}
For KernelBench \cite{ouyang2025kernelbenchllmswriteefficient} , to decompose the main metric, we adopt the following metrics 1) Compilation Success, 2) Runtime Success, 3) Correctness, 4) Fast 1 and 5) Fast 2.
% 1) LLM Format Output, 2) Syntax Check, 3) Compilation Check, 4) Model Loading Check, 5) runtime check, 6) correctness check, 7) fast 1 metric and 8) fast 2 metric. 
For more details see Appendix \ref{app:metrics}.

This structure follows the requirement presented in Section \ref{sec:formal_def_multi_stage} that $\mathcal{V}_i(y,\hat{y}) \geq \mathcal{V}_j(y,\hat{y})$ whenever $i<j$ and $\mathcal{V}_N = \mathcal{V}$.

In our case each $\mathcal{V}_i$ is a binary variable and follows a strict order in that if a preceding metric is false all subsequent metrics will be false and therefore 0. The thresholds required for our algorithm are also set trivially to 1 for each metric.

\subsubsection{Inductive Biases for Each Metric}
Finally, we choose the inductive biases to be extensions of the prompt to the LLM that explicitly bias the LLM to produce solutions in the given failing metric, e.g. compilation, or runtime, etc., see Appendix \ref{app:inductive_bias} for the complete list of inductive biases.

\subsection{Baselines}
In terms of baselines we use the current best in class test time scaling method based on iterative refinement \cite{ouyang2025kernelbenchllmswriteefficient, baronio2026kevin}.
% ; as well as RL training on the task \cite{baronio2026kevin}. 
In our study we take a strong iterative refinement baseline, based on the current best in class methods \cite{ouyang2025kernelbenchllmswriteefficient, baronio2026kevin}, which represents the data-driven approach. We test it against our method by keeping everything fixed and adding only the inductive bias.

Concrete, implementation details are given in Appendix \ref{app:implementation}.

%TODO: 

% \begin{enumerate}
%     \item LLM Format Output (\textbf{new})- which checks if the output of the LLM is in a valid format
%     \item Syntax Check (\textbf{new})- which checks whether basic syntax of the code is correct
%     \item Compilation Check - which checks if the code compiles successfully.
%     \item Model Loading Check (\textbf{new}) - which checks if the new operand can be loaded successfully
%     \item Runtime Check (\textbf{new}) - which checks whether the operand can be run successfully
%     \item Correctness Check - which checks whether the operand produces the same output (within a floating point error tolerance) as the PyTorch operand
%     \item Fast 1 - which measures the ratio of operands (Kernels) that succeeded in the Correctness Check, but also have a speed-up of at least 1 (i.e. they are not slower than PyTorch)
%     \item Fast 2 - which measures the same as Fast 1, but with a minimum speed-up of 2 against PyTorch.
% \end{enumerate}

% \subsection{Reproducibility}
% Maybe we want to say something about reprodcibility? 
% %TODO: 

\section{Results}
\begin{figure*}[ht]
  \vskip 0.2in
  \begin{center}
    \centerline{\includegraphics[width=\linewidth]{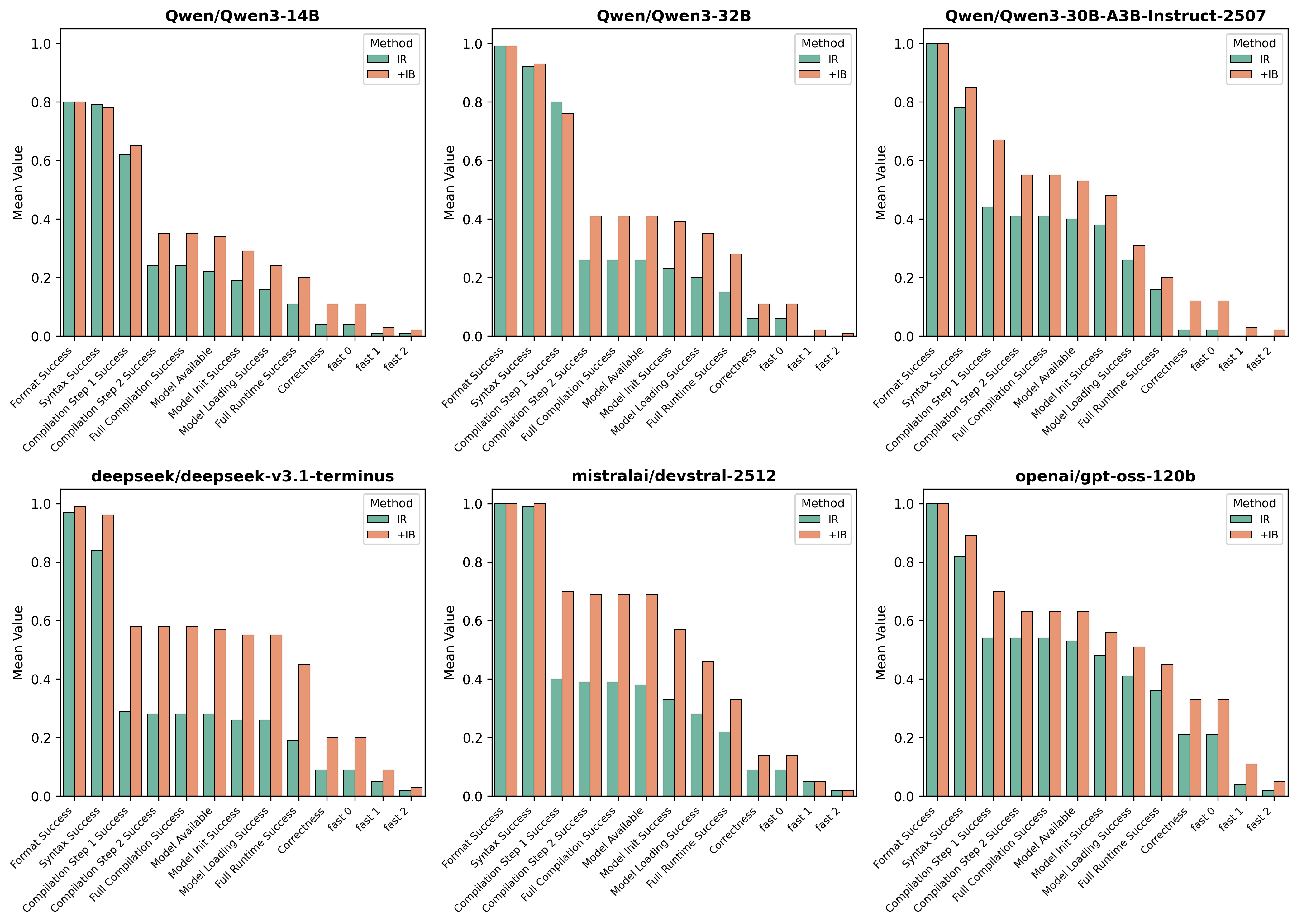}}
    \caption{
      % Results for Qwen-Coder, for iterative refinement vs. our multi-stage iterative refinement.
      Results across various models from Table \ref{tab:main_results}. X-axis represents the various metrics. Y-axis represents the normalized score from the 100 samples. IR (green) represents the iterative refinement approach; IB (orange) represents our inductive bias based iterative refinement.
    }
    \label{fig:main_results}
  \end{center}
\end{figure*}
We can see our results in Figure \ref{fig:main_results} and \ref{app:additional_results} Table \ref{tab:main_results}. The x-axis represents the various metrics we used; while the y-axis represents the aggregate score.

\paragraph{Uniform improvement from inductive bias}
For every model and important metrics like compilation, runtime and correctness, the inductive bias prompt shows significant improvement over the iterative refinement baseline. No metric degrades. This consistency across 4 architecturally diverse models: Qwen, DeepSeek, Mistral and GPT. This strongly suggests that the advantage of inductive bias is not an artifact of a particular model family.

\paragraph{Magnitude of gains}
The largest absolute improvements are observed in compilation success, with deltas ranging being up to $+0.31$ for Mistral's Devstral-2512. Runtime success improves by up to $+0.17$, and correctness by up to $+0.12$ for gpt-oss. Relative improvements are often substantial: for example, Qwen3-14B shows a $2$x relative increase in correctness, from $0.04$ to $0.11$. Even the strongest iterative refinement baseline, GPT-OSS-120B with correctness $0.21$ is outperformed by its own inductive bias variant $0.33$, indicating that inductive biases benefit both weak and state‑of‑the‑art models.

\paragraph{Correctness remains challenging.}
Despite consistent improvements, absolute correctness under multi stage search does not exceed $0.33$ for any model. This reflects the inherent difficulty of the GPU programming task, which highly sophisticated low level parallel code. Inductive biases provide a meaningful but not yet sufficient advantage.

\subsection{Model Specific Observations}

\begin{itemize}
    % \item \textbf{DeepSeek-Coder-V2-Lite-Instruct} achieves the highest iterative refinement compilation success ($0.61$) and shows no further gain from multi stage search on that metric ($0.61$). This ceiling effect suggests that this model may already incorporate implicit inductive biases from its code specific pretraining, leaving little room for additional prompt‑level constraints.
    \item \textbf{DeepSeek V3.1 Terminus} and \textbf{Mistral Devstral-2512} exhibit the largest improvements from the inductive biases, particularly in compilation ($+0.22$ and $+0.29$, respectively). These models appear to benefit most from explicit solution space constraints.
    \item \textbf{Qwen3-32B} under iterative refinement performs poorly (compilation $0.26$) but rises to a competitive $0.41$ with inductive biases, nearly matching the iterative refinement performance of much larger models. This further highlights how inductive biases alleviate the `smaller' data regime.
    \item \textbf{GPT-oss-120b} achieves on average the best scores across most metrics and especially fast 1 and fast 2. Interestingly, inductive biases are very effective with this model for all metrics.
    \item The performance metrics \texttt{fast1} and \texttt{fast2} show only a small direction of change; some models see small increases with inductive biases, only Qwen3-30B-A3B has any statistical significance. Again this indicates that this field of designing effective inductive biases for harder fields is still open.
\end{itemize}

\begin{table*}[htbp]
\centering
\caption{Main results table summarizing the scores. All experiments were conducted on the KernelBench GPU programming task with a fixed test set of size $N = 100$ problems from level~1. IR = Iterative Refinement; IB = Inductive Bias. Reported values are mean $\pm$ standard deviation over 3 runs. Green bold values indicate statistically significant improvement of +IB over IR (Welch's $t$-test, one‑tailed, $p < 0.05$). \textbf{Bold} = best overall, \underline{underline} = second best (ties allowed). COMP = Compilation Success, RUN = Runtime Success, CORR = Correctness.}
\label{tab:main_results}
\begin{tabular}{llccccc}
\toprule
\textbf{Model} & \textbf{Method} & \textbf{COMP} & \textbf{RUN} & \textbf{CORR} & \textbf{Fast1} & \textbf{Fast2} \\
\midrule
Qwen3-14B & IR & 0.22\stdpm{0.06} & 0.19\stdpm{0.03} & 0.04\stdpm{0.02} & 0.01\stdpm{0.01} & 0.01\stdpm{0.01} \\
Qwen3-14B & +IB & \sigimprove{0.34}\stdpm{0.04} & \sigimprove{0.29}\stdpm{0.05} & \sigimprove{0.11}\stdpm{0.01} & \sigimprove{0.03}\stdpm{0.01} & 0.02\stdpm{0.01} \\
\hline
Qwen3-32B & IR & 0.26\stdpm{0.05} & 0.23\stdpm{0.05} & 0.06\stdpm{0.01} & 0.00\stdpm{0.01} & 0.00\stdpm{0.01} \\
Qwen3-32B & +IB & \sigimprove{0.41}\stdpm{0.02} & \sigimprove{0.39}\stdpm{0.02} & \sigimprove{0.11}\stdpm{0.02} & 0.02\stdpm{0.02} & 0.01\stdpm{0.01} \\
\hline
Qwen3-30B-A3B & IR & 0.40\stdpm{0.00} & 0.38\stdpm{0.00} & 0.02\stdpm{0.00} & 0.00\stdpm{0.00} & 0.00\stdpm{0.00} \\
Qwen3-30B-A3B & +IB & \sigimprove{0.53}\stdpm{0.00} & \sigimprove{0.48}\stdpm{0.00} & \sigimprove{0.12}\stdpm{0.00} & \sigimprove{0.03}\stdpm{0.00} & \sigimprove{0.02}\stdpm{0.00} \\
\hline
deepseek-v3.1 & IR & 0.28\stdpm{0.01} & 0.26\stdpm{0.02} & 0.09\stdpm{0.01} & 0.05\stdpm{0.01} & 0.02\stdpm{0.02} \\
deepseek-v3.1 & +IB & \sigimprove{0.57}\stdpm{0.08} & \sigimprove{0.55}\stdpm{0.07} & \sigimprove{0.20}\stdpm{0.03} & \underline{\sigimprove{0.09}\stdpm{0.02}} & \underline{0.03}\stdpm{0.01} \\
\hline
devstral-2512 & IR & 0.38\stdpm{0.02} & 0.33\stdpm{0.02} & 0.09\stdpm{0.01} & 0.05\stdpm{0.02} & 0.02\stdpm{0.01} \\
devstral-2512 & +IB & \sigimprove{0.69}\stdpm{0.02} & \sigimprove{0.57}\stdpm{0.01} & \sigimprove{0.14}\stdpm{0.01} & 0.05\stdpm{0.01} & 0.02\stdpm{0.01} \\
\hline
gpt-oss-120b & IR & 0.53\stdpm{0.03} & 0.48\stdpm{0.05} & \underline{0.21}\stdpm{0.02} & 0.04\stdpm{0.01} & 0.02\stdpm{0.02} \\
gpt-oss-120b & +IB & \underline{\sigimprove{0.63}\stdpm{0.05}} & \underline{0.56}\stdpm{0.07} & \sigimprove{0.33}\stdpm{0.05} & \textbf{0.11}\stdpm{0.06} & \textbf{0.05}\stdpm{0.03} \\
\bottomrule
\end{tabular}

\end{table*}

\subsection{Control Study: Effect of Additional Coding Pre-training Data}
In order to test the hypothesis that inductive biases are truly effective in the low data regime, we construct a small controlled study where we test the `coding' models in two LLM families, Qwen and DeepSeek. This controlled experiment is in line with the hypothesis that more data reduces the effect of inductive biases. Conversely, this also affirms that reducing data increases the role of inductive biases. Table \ref{tab:coder_comparison} and Table \ref{tab:additional_results_appendix}, clearly show that the effect of inductive biases drops as more data is added.

\begin{table*}[htbp]
\centering
\caption{Results for the `Coder' models in the Qwen and DeepSeek family, it can be seen that the inductive bias method has only marginal improvement in the case of further pre-training. See Table \ref{tab:main_results} for details.}
\label{tab:coder_comparison}
\begin{tabular}{llccccc}
\toprule
\textbf{Model} & \textbf{Method} & \textbf{COMP} & \textbf{RUN} & \textbf{CORR} & \textbf{Fast1} & \textbf{Fast2} \\
\midrule
Qwen3-30B-A3B & IR & 0.40\stdpm{0.00} & 0.38\stdpm{0.00} & 0.02\stdpm{0.00} & 0.00\stdpm{0.00} & \underline{0.00}\stdpm{0.00} \\
Qwen3-30B-A3B & +IB & \sigimprove{0.53}\stdpm{0.00} & \underline{\sigimprove{0.48}\stdpm{0.00}} & \sigimprove{0.12}\stdpm{0.00} & \underline{\sigimprove{0.03}\stdpm{0.00}} & \sigimprove{0.02}\stdpm{0.00} \\
\hline
Qwen3-Coder-30B-A3B & IR & \underline{0.60}\stdpm{0.00} & \textbf{0.59}\stdpm{0.00} & \underline{0.22}\stdpm{0.00} & \textbf{0.05}\stdpm{0.00} & \textbf{0.02}\stdpm{0.00} \\
Qwen3-Coder-30B-A3B & +IB & \sigimprove{0.63}\stdpm{0.00} & \textbf{0.59}\stdpm{0.00} & \sigimprove{0.23}\stdpm{0.00} & 0.01\stdpm{0.00} & \underline{0.00}\stdpm{0.00} \\
\hline
DeepSeek-Coder & IR & 0.58\stdpm{0.00} & 0.50\stdpm{0.00} & 0.18\stdpm{0.00} & 0.03\stdpm{0.00} & 0.00\stdpm{0.00} \\
DeepSeek-Coder & +IB & \sigimprove{0.59}\stdpm{0.00} & \sigimprove{0.54}\stdpm{0.00} & \sigimprove{0.20}\stdpm{0.00} & \sigimprove{0.05}\stdpm{0.00} & \underline{\sigimprove{0.03}\stdpm{0.00}} \\
\bottomrule
\end{tabular}
\end{table*}

\section{Conclusion}
We explored a new scaling dimension for AI: the duration of the \textit{Feedback Information Loop} (FIL). Nowadays the prevailing ML paradigm leverages vast amounts of near instantaneous data to drive improvements. We argued that many emerging applications in science, engineering, and the physical world are characterized by inherently slow FILs ranging from hours to weeks. In such regimes purely data driven approaches that have dominated recent AI successes become impractical.

We use KernelBench as a slow FIL task to investigate this hypothesis. Specifically, KernelBench has slow reward signals due to its non trivial verification. The current state-of-the-art relies on iterative refinement and achieves only modest success, underscoring the limitations of short FIL approaches even in minute scale tasks.

As a first step toward addressing these constraints, we proposed an inductive bias based method that decomposes the verification process into an ordered hierarchy of binary metrics and injects stage specific inductive biases via targeted prompting. This approach leverages structured prior knowledge to produce better predictions.

Our results demonstrate that incorporating such inductive biases, consistently outperforms existing iterative refinement approaches across a range of open and frontier models. These serves as an initial validation that, as FIL duration grows, explicit inductive biases can once again become an important component for ML algorithms and complementing or even surpassing purely data driven scaling.

Looking forward, we hope this work encourages the community to systematically study and develop methods for the emerging problem of slow FIL AI.

% As we push toward applications with FILs of days or weeks, the principles explored here ordered decomposition, targeted inductive guidance, and efficient use of sparse feedback will likely become foundational. 
% \section*{Impact Statement}

% This paper presents work whose goal is to advance the field of Machine
% Learning. There are many potential societal consequences of our work, none
% which we feel must be specifically highlighted here.

\section*{Acknowledgements}
This work has been supported by the German Federal Ministry of Research, Technology and Space and the Hessian Ministry of Higher Education, Research, Science and the Arts within their joint support of the National Research Center for Applied Cybersecurity ATHENE.

% \section*{References}

\bibliography{bibliography}
\bibliographystyle{myacm}

%%%%%%%%%%%%%%%%%%%%%%%%%%%%%%%%%%%%%%%%%%%%%%%%%%%%%%%%%%%%

\appendix

% \section{Motivation \& Background}
\section{Background}
\subsection{Learning with delayed and sparse feedback}
\label{subsec:delayed_feedback}
% TODO:

\begin{figure}[ht!]
  \vskip 0.2in
  \begin{center}
    % \centerline{\includegraphics[width=0.9\columnwidth]{images/delay_vs_fil.png}} %two column setting
    \includegraphics[width=0.45\linewidth]{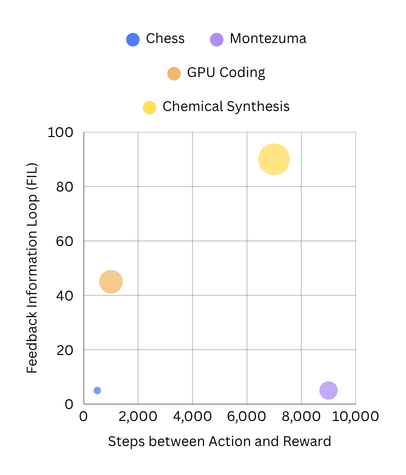}
    \caption{
      Schematic illustration of the difference between delay in reward due to number of steps in the environment versus delay in reward due to increased FIL, showing that these are orthogonal properties.
    }
    \label{fig:delay_vs_FIL}
  \end{center}
\end{figure}

% \todo{This subsection needs clearer wording. Why do we think that algorithmic way is not the right way and how our method solves this.}
The challenge of learning from delayed feedback has been a persistent concern across multiple subfields of machine learning, especially in reinforcement learning (RL). The problem is often referred to as a sparse rewards setting. This both degrades sample efficiency and often prevents meaningful asymptotic solutions \citep{NEURIPS2022_266c0f19_Sparse_rewards, burda2018exploration}. In the sparse or delayed reward setting, the delay can be formalized as the number of environment steps between an action and the reward for that action, the goal is to learn a policy that performs well \emph{despite} this delay. This setting is orthogonal to the one presented in this paper as can be seen in Figure \ref{fig:delay_vs_FIL}, since both delay due to the environment dynamics and due to FIL are independent.

Existing work has approached this problem of delayed environment dynamics 
% primarily as an \emph{algorithmic} challenge, such as 
via curiosity driven algorithms \cite{burda2018exploration}, reward shaping \cite{NEURIPS2022_266c0f19_Sparse_rewards, 10.5555/3454287.3455502NeuripsRudder2019} and principled statistics algorithms, such as Bayes Exploration in RL \cite{osband_bayes_rl_10.5555/3157382.3157548}
% or world models and model-based RL such as DreamerV3 \cite{Hafner2025DreamerV3}. 
While such approaches represents a meaningful advance, they represent an orthogonal approach to the one we discuss in this study, as the existing approaches are still data driven.
% % it operates under an orthogonal assumption than the one we examine in this work. 
% In the sparse or delayed reward setting, the delay is
% % , the delay is a fixed property of the 
% % \emph{action-observation loop}, 
% the number of environment steps between an action and the reward for that action, the goal is to learn a policy that performs well \emph{despite} this delay. 
% The delay itself is not a resource constraint on data collection; it is a property of the environment dynamics. 
Our setting, by contrast, focuses on regimes where 
% the delay is a property of 
the \emph{verification process itself}
% the time required to obtain a reward signal after generating a candidate solution. In this regime, the delay directly 
determines the maximum feasible data scale, transforming what has traditionally been an algorithmic challenge into a fundamental data bottleneck.

\subsection{Approximate feedback loop}
\label{subsec:approximate_feedback}

A parallel line of research seeks to circumvent long feedback delays by learning \emph{surrogate models} that approximate it. Model based RL and the more recent work on  \textbf{world models} learn to predict the environment dynamic and represent the most systematic instantiation of this approach \citep{NEURIPS2018_2de5d166_world_models_2018, pmlr-v267-zhang25j_world_model_theory, Balestriero2025LeJEPAPA, rozanov2026multitaskllmsbugclassification}. By enabling an agent to ``imagine'' the outcomes of potential actions without executing them in the real environment, world models can dramatically reduce the need for real world interaction.

The development of world models has progressed through several architectural generations. PlaNet \citep{pmlr-v97-hafner19a_learning_latent_dynamics} introduced learning latent dynamics from pixels for planning. The Dreamer family of algorithms \citep{Dreamerv1_hafner, Hafner2025DreamerV3} extended this paradigm to scalable reinforcement learning, achieving state-of-the-art performance on Atari and continuous control benchmarks by learning behaviors entirely within latent imagination. Recent work has expanded world models to incorporate object-centric representations \citep{wu2023slotformer}, which improve compositional generalization and sample efficiency in visually challenging tasks. Transformers have been integrated into world models \citep{chen2021transdreamer, pmlr-v235-micheli24a_transformer_world_model} to capture longer-range dependencies. The application domains for world models now span robotics, autonomous driving, scientific discovery, coding and many more.

While these approaches are successful for short FIL, \textbf{the world model approach faces fundamental limitations that are directly relevant to our thesis.} Firstly, the \textbf{simulation to real gap}: policies trained entirely within learned simulators often fail when deployed in the real world due to distribution shift between the model's training distribution and the true environment dynamics. This challenge is extensively documented in different fields, including robotics, natural and social sciences. Recent surveys explicitly identify ``Physical and Natural Sciences'' and ``Social Science and Socioeconomic Systems'' as an area where world models exhibit a significant struggle \citep{10.1145/3746449_world_model_survey}. Second, and more fundamentally, the world model itself must be trained on real-world data. Learning an accurate simulator of chemical synthesis, aerodynamic design, or GPU kernel performance still requires collecting verification data from the real world. The world model approach does \emph{not} eliminate the Feedback Information Loop. If the FIL for a single verification is one week, collecting the dataset to train a reliable world model may require thousands or millions of such verifications, which is precisely the infeasible regime we identify.

Thus, while world models represent a powerful complementary technique, they do not resolve the underlying problem that some environments will require real verification process that creates a concrete feedback bottleneck.

\subsection{Existing work on inductive biases and using expert knowledge and constraints}
\label{subsec:inductive_biases_others}

Inductive biases are often characterized as structural priors embedded within machine learning models guide generalization \cite{pmlr-v267-wilson25a_soft_inductive_bias, subramaniam2025training_training_untrainable}.
Notably, work on diffusion models has revealed that locally linear UNet based denoisers possess specific inductive biases that enable generalization under constrained conditions \cite{an2025on_inductive_biases_diffusion}. Additionally, studies demonstrate that language models require explicit inductive biases to perform certain reasoning tasks like recursive counting, challenging the notion that scale alone suffices \cite{chang2025language_inductive_bias_for_counting}. 

Additionally, there have been several works studying the systematic injection of knowledge into machine learning systems \cite{9429985_survey_of_knowledge_in_models, lauscher-etal-2020-common, rozanov2025stateactenhancingllmbase, ibrahim2025finetuningragimprovingllm}, termed \textit{informed machine learning}, which posits four stages at which prior knowledge can be integrated into a machine learning pipeline: 1) the training data, 2) the hypothesis set, 3) the learning algorithm, and 4) the final hypothesis choice.
Existing work of \textit{informed machine learning} operates generally under the assumption of enhancing existing solutions with expert knowledge.
% , as opposed to the fundamental limitation that such data is not available. 

In our work we enhance the concept of \textit{informed} machine learning to the setting where the FIL duration becomes expensive and therefore posits fundamental limits to the data collection process.

\section{Implementation Details}
\label{app:implementation}

\begin{figure}
    \centering
    \includegraphics[width=0.9\linewidth]{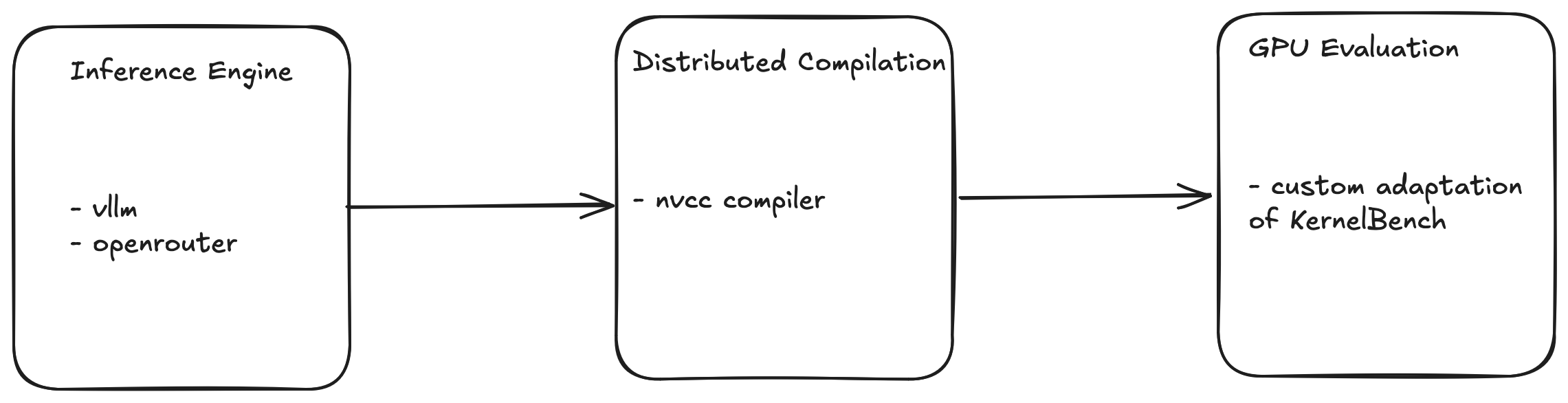}
    \caption{Data flow during the evaluation of a single sample.}
    \label{fig:evaluation_flow}
\end{figure}

Our evaluation code and prompts will be open-sourced upon publication. Generally, we use our own adaptation of the original KernelBench evaluation suite and made it more robust. Our flow consists of three parts: 1) Inference either via API or vLLM \cite{vllm_10.1145/3600006.3613165}; 2) Distributed Compilation using Nvidia's NVCC compiler; 3) Evaluation using our own custom evaluation script; that overcomes existing issues with KernelBench\footnote{The following issue crashes the original benchmark: \url{https://github.com/ScalingIntelligence/KernelBench/issues/76}.}. Figure \ref{fig:evaluation_flow} demonstrates our data flow schematically.

\subsection{Models}
In our paper we aim at reproducibility, therefore we priortize results and experiments using open models. We investigate four model families, Qwen, GPT, Mistral and DeepSeek, specifically see Table \ref{tab:models}.
% family of models, with sizes ranging from 
% 14B and 32B. Additionally we evaluate across frontier models from Mistral, OpenAI, and DeepSeek; specifically, Devstral-0513, gpt-oss-120B, and V3.1 and V2-Coder, respectively.

\begin{table}
    \centering
    \caption{The models we used in our work.}
    \begin{tabular}{l|l|l}
        \toprule
         Model Type&  Size&  Inference Provider\\
         \hline
         Qwen3-4B&  14B&  vLLM \\
         Qwen3-8B&  14B&  vLLM \\
         Qwen3-14B&  14B&  vLLM \\
         Qwen3-32B&  32B&  vLLM \\
         Qwen3-30B-A3B-Instruct&  30B&  vLLM \\
         Qwen3-Coder-30B-A3B-Instruct&  30B&  vLLM \\
         DeepSeek-V2-Coder&  16B&  vLLM \\
         DeepSeek-V3.1&  671B&  OpenRouter \\
         Devstral-0512 (Devstral 2)&  123B&  OpenRouter \\
 Gpt-oss-120B& 120B& OpenRouter\\
          \bottomrule
    \end{tabular}
    \label{tab:models}
\end{table}

All hyper-parameters were either default values of OpenRouter and vLLM respectively or set based on Table \ref{tab:hyper_params}.

\begin{table}[h!]
    \centering
    \caption{Hyper-parameters that we set fixed for all runs.}
    \begin{tabular}{c|c}
    \toprule
        Hyper-parameter & Value\\
        \hline
         Temperature & 0.6 \\
                 \hline
        Max Model Len & 32000\\
                 \hline
        Max Tokens & 8000\\
                 \hline
         Dtype & Automatic (i.e. default for each model) \\
                 \bottomrule
    \end{tabular}
    \label{tab:hyper_params}
\end{table}

% \subsection{Verification Functions}
% TBC...

\subsection{Computational Infrastructure}
All experiments were run using A40 GPUs;
% provided by MBZUAI; 
or A100 GPUs
% provided by UKP Lab. 
Overall, more than 500 GPU hours were spent on the project. 
Additionally, OpenRouter credits were used.
% provided by MBZUAI were used.

\subsection{Licenses}
\paragraph{LLM Models} all models have open-weight licenses that we used. We were allowed to use them. The work was cited.

\paragraph{Dataset} KernelBench, is licensed under MIT, so free to use. The work was cited.

\paragraph{Software} All software is opensource and open for use, only standard libraries were used or own software. All relevant software was cited.
% \newpage
\section{Inductive bias search}
\label{app:multi_stage_search}

\paragraph{Deriving the Probabilistic Relation}Let $p(y)$ denote the base sampling distribution induced by the underlying model. Inductive bias search modifies this distribution through conditioning on intermediate constraints. At stage $i$, the effective sampling distribution is:
\[p_i(y) = p(y \mid y \in \mathcal{Y}_{i-1})\]

Define the stage-wise success probability:
\[p_i = \mathbb{P}_{y \sim p_{i-1}}(V_i(y) = 1)\]
i.e., the probability that a sample drawn at stage $i-1$ satisfies the constraint introduced at stage $i$. The probability of sampling a solution that satisfies all stages can be decomposed as:
\[p_{\mathrm{final}} = \prod_{i=1}^N p_i\]

The monotonicity of stage-wise success probabilities $p_i$ does not hold in general, as each $p_i$ is defined with respect to a different conditional distribution. Instead, we consider the conditional success probability of the final objective:
\[
\tilde{p}_i = \mathbb{P}(V_N(y) = 1 \mid y \in \mathcal{Y}_i)
\]
Under effective decomposition, we expect:
\[
\tilde{p}_0 \le \tilde{p}_1 \le \dots \le \tilde{p}_N = 1
\]
that is, successive constraints increase the likelihood that a candidate satisfies the final objective.

\subsection{Inductive bias algorithm}

% OK this is now a working algorithm...
\begin{algorithm}[h!]
\caption{Inductive bias search}
\label{alg:multi_stage_search}
\begin{algorithmic}[1]
\REQUIRE Input $x\in\mathcal{X}$, output $y\in\mathcal{Y}$, metric $\mathcal{V}$, iteration budget M, (existing) $LLM(prompt,x)$
\ENSURE Solution $y^*$, ordered decomposition $\left \{ \mathcal{V}_i \right \}_{i=1}^N$, thresholds $\mathcal{T}=\{t_1,...,t_N\}$,  inductive biases $\mathcal{B}=\{inductive\_bias_1,...,inductive\_bias_N\}$
% for $i = 1, 2, \dots, n$
% \STATE Initialize token pointer $p \gets 1$, line pointer $q \gets 1$
\STATE Human expert proposes and creates ordered decomposition of metric $\mathcal{V}$ alongside appropriate thresholds and the inductive biases in N stages. 
\STATE Initialize $\hat{y} \gets LLM(prompt,x)$
\FOR{$i = 1, 2, \dots, M$} 
    % \STATE $current\_stage \gets 1$
    \FOR{$j = 1, 2, \dots, N$} 
        \STATE $current\_score \gets \mathcal{V}_j(y,\hat{y})$
        \IF{$current\_score < t_j$}
            % \STATE $current\_stage \gets j$
            \STATE $\hat{y} \gets LLM(prompt+inductive\_bias_j, x)$
            \STATE \textbf{break}
        \ENDIF
    \ENDFOR
    % \STATE $\hat{y} \gets LLM(prompt+inductive\_bias_j, x)$
\ENDFOR
\STATE \textbf{return} $\hat{y}$
\end{algorithmic}
\end{algorithm}

\subsection{Comparison of Classical Search vs. Multi-Stage Search}
\begin{figure*}[ht]
  % \vskip 0.2in
  \begin{center}
 % \centerline{\includegraphics[width=0.5\linewidth]{images/classical_search.png}}
 \centerline{\includegraphics[width=0.7\linewidth]{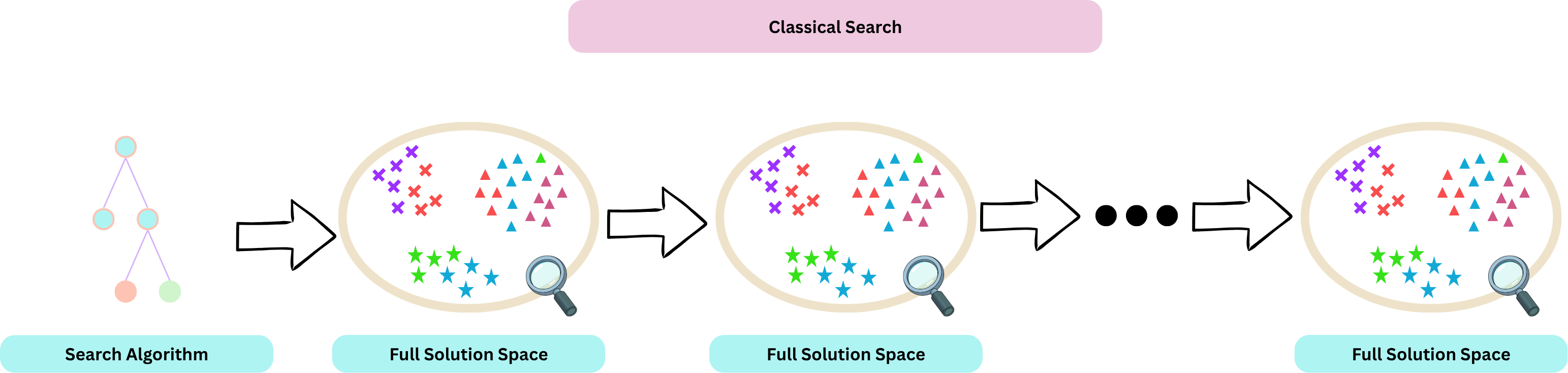}}
    \caption{
      Classical Solution Search
    }
    \label{fig:classic_search}
  \end{center}
\end{figure*}

\begin{figure*}[ht]
  % \vskip 0.2in
  \begin{center}
    %width=0.9
    \centerline{\includegraphics[width=0.7\linewidth]{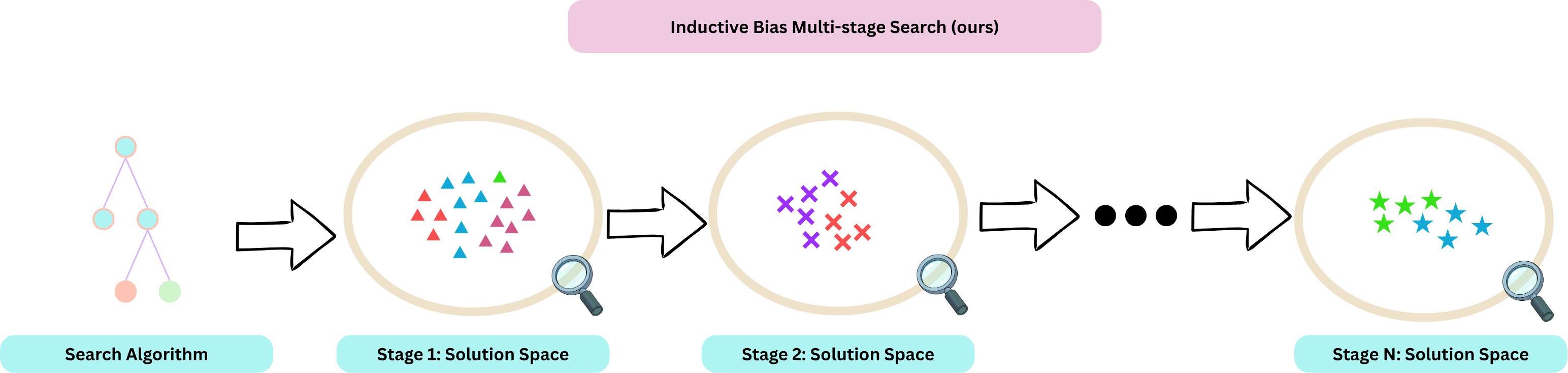}}
    \caption{
      Inductive bias search (Ours)
    }
    \label{fig:multi_stage_search}
  \end{center}
\end{figure*}

% \newpage
\section{Metrics used for KernelBench}
\label {app:metrics}

% \begin{table}[h]
% \centering
% \caption{Metrics for KernelBench evaluation. Some are from \cite{ouyang2025kernelbenchllmswriteefficient} while other metrics are newly introduced.}
% \begin{tabular}{p{4cm} p{8cm} c}
% \toprule
% \textbf{Metric} & \textbf{Description} & \textbf{New} \\
% \midrule
% LLM Format Output & Checks if the output of the LLM is in a valid format & Yes \\
% \midrule
% Syntax Check & Checks whether basic syntax of the code is correct & Yes \\
% \midrule
% Compilation Check & Checks if the code compiles successfully & No \\
% \midrule
% Model Loading Check & Checks if the new operand can be loaded successfully & Yes \\
% \midrule
% Runtime Check & Checks whether the operand can be run successfully & Yes \\
% \midrule
% Correctness Check & Checks whether the operand produces the same output (within a floating point error tolerance) as the PyTorch operand & No \\
% \midrule
% Fast 1 & Measures the ratio of operands (Kernels) that succeeded in the Correctness Check, but also have a speed-up of at least 1 (i.e. they are not slower than PyTorch) & No \\
% \midrule
% Fast 2 & Measures the same as Fast 1, but with a minimum speed-up of 2 against PyTorch & No \\
% \bottomrule
% \end{tabular}
% \label{tab:kernelbench_metrics}
% \end{table}

\begin{table}[h]
\centering
\caption{Metrics for KernelBench evaluation. Some are from \cite{ouyang2025kernelbenchllmswriteefficient} while other metrics are newly introduced.}
\begin{tabular}{p{3.5cm} p{8cm} c}
\toprule
\textbf{Metric} & \textbf{Description} & \textbf{New} \\
\toprule
Format Success& Checks if the LLM output is correct& Yes\\
\midrule
Syntax Success& Checks whether the syntax of the code is correct& Yes \\
\midrule
Compilation Step 1 Success& Does a simple compilation of the file, but does not try to load the file itself.& Yes\\
\midrule
Compilation Step 2 Success& Compiles and loads the file& Yes\\
\midrule
Full Compilation Success& Compiles, loads the file and loads the new model.& No \\
\midrule
 Model Available& A second attempt at compiling, loading the file and loading the new model from scratch.&Yes\\
 \midrule
 Model Init Success& Attempts to initalize the model class, without any weight transfer.&Yes\\
 \midrule
 Model Loading Success& Attempts to load the new model fully.&Yes\\
 \midrule
 Full Runtime Success& Runs the model and compares against the reference model.&No\\
 \midrule
 Correctness& Checks whether the output is correct.&No\\
 \midrule
 Fast 1& Counts how many solutions are both correct and at least 1x faster than the reference solution (i.e. same speed)&No\\
 \midrule
 Fast 2& Same as Fast 1, but at least 2x faster than the reference solution.&No\\
 \bottomrule
\end{tabular}
\label{tab:kernelbench_metrics}
\end{table}

\newpage
\section{Inductive Biases for KernelBench}
\label {app:inductive_bias}

\begin{table}[h]
\centering
\caption{The below are the metrics we choose for the ordered decomposition; as well as the inductive biases we chose for our method.}
\begin{tabular}{p{1.5cm} p{11cm}}
\toprule
Compilation Success & 
\texttt{``Unfortunately, your previous answer failed to compile. Think carefully about how to solve the compilation issues. Here is the error message: $\{error\_message\}$''
``Remember your key objective now is to debug the code and arrive at a correct CUDA Implementation.''
} 
\\
\midrule
Runtime Success & \texttt{``Your previous answer compiled successfully but had runtime errors. Think carefully how to solve the runtime issues. Here is the error message: $\{error\_message\}$''
``Remember your key objective now is to debug the code and arrive at a correct CUDA Implementation.''
}  \\
\midrule
Correctness Success & \texttt{``Your previous answered compiled and ran without problems. However, your previous code produced the wrong answer. I.e. either the shape or the numbers were wrong. Think carefully how to make your code produce the correct code. Here is the error message: $\{error\_message\}$''
``Remember your key objective now is to make sure your code produces the correct answer.''
} 
\\
\midrule
Fast 1 &
\texttt{``Your previous answer was correct but can be made faster. Think carefully how to make the solution faster. Here is the speedup you achieved relative to the baseline: $\{score\}$times faster. ''
``Remember your key objective now is to make the code run faster.''} 
\\
\midrule
Fast 2 &
\texttt{``Your previous answer was correct but can be made faster. Think carefully how to make the solution faster. Here is the speedup you achieved relative to the baseline: $\{score\}$times faster. ''
``Remember your key objective now is to make the code run faster.''} 
\\
\bottomrule
\end{tabular}
\label{tab:kernelbench_inductive_bias}
\end{table}

\newpage
\section{Feedback Information Loop Tables}
\label{app:FIL_tables}

\begin{table}[htbp]
\centering
\caption{Selected Tasks with Prolonged Feedback Information Loop.}
\label{tab:data_needs_2_detailed}
% \begin{tabular}{p{3.8cm} p{3.8cm} p{3.2cm}}
\begin{tabular}{p{4cm} p{4.5cm} p{3.4cm}}
\toprule
\textbf{Domain} &  \textbf{Type of Evaluation} &  \textbf{Feedback Information Loop (FIL)} \\
\midrule
GPU Kernel Programming \cite{ouyang2025kernelbenchllmswriteefficient}  & Code Compilation and Performance Evaluation & $\sim$2 - 30 minutes\\
\hline
Machine Learning Engineering \cite{chan2025mlebench}   & Datascience and Machine Learning Challenges  & $\sim$2 minutes - 1 hour \\
\hline
OS Kernel Engineering  & Compiling new OS Kernels and checking if a specific metric has improved (performance, memory, security) & $\sim$ 5 minutes \cite{borges2025linuxkernelconfigurationsscale} \\
\hline
Real Chemical Synthesis   & Checking if a proposed synthesis plan yields a given substance & $\sim <$ 1 hour - several days - several weeks \cite{Volk2023_alpha_flow_chemical_synthesis} \\
\hline
Automatic Large Language Model Research & Checking whether a given LLM recipe (training data and training pipeline) yields an improved LLM & $\sim <$ 1 day - several days - several months \cite{Pretraining_runs_100k_100days} \\
% \hline
\bottomrule
\end{tabular}
\end{table}

\newpage
\section{Additional Results}
\label{app:additional_results}

\begin{table*}[htbp]
\centering
\caption{Additional results table summarizing the scores of all tested models. All experiments were conducted on the KernelBench GPU programming task with a fixed test set of size $N = 100$ problems from level~1. IR = Iterative Refinement; IB = Inductive Bias. Reported values are mean $\pm$ standard deviation over 3 runs. Green bold values indicate statistically significant improvement of +IB over IR (Welch's $t$-test, one‑tailed, $p < 0.05$). \textbf{Bold} = best overall, \underline{underline} = second best (ties allowed). COMP = Compilation Success, RUN = Runtime Success, CORR = Correctness.}
\label{tab:additional_results_appendix}
\begin{tabular}{llccccc}
\toprule
\textbf{Model} & \textbf{Method} & \textbf{COMP} & \textbf{RUN} & \textbf{CORR} & \textbf{Fast1} & \textbf{Fast2} \\
\midrule
Qwen3-4B & IR & 0.52\stdpm{0.00} & 0.50\stdpm{0.00} & 0.05\stdpm{0.00} & 0.02\stdpm{0.00} & 0.00\stdpm{0.00} \\
Qwen3-4B & +IB & \sigimprove{0.54}\stdpm{0.01} & \sigimprove{0.52}\stdpm{0.01} & 0.02\stdpm{0.00} & 0.02\stdpm{0.00} & 0.00\stdpm{0.00} \\
\hline
Qwen3-8B & IR & 0.13\stdpm{0.00} & 0.11\stdpm{0.00} & 0.03\stdpm{0.00} & 0.01\stdpm{0.00} & 0.00\stdpm{0.00} \\
Qwen3-8B & +IB & 0.11\stdpm{0.00} & 0.09\stdpm{0.00} & \sigimprove{0.05}\stdpm{0.00} & \sigimprove{0.03}\stdpm{0.00} & \sigimprove{0.02}\stdpm{0.00} \\
\hline
Qwen3-14B & IR & 0.22\stdpm{0.06} & 0.19\stdpm{0.03} & 0.04\stdpm{0.02} & 0.01\stdpm{0.01} & 0.01\stdpm{0.01} \\
Qwen3-14B & +IB & \sigimprove{0.34}\stdpm{0.04} & \sigimprove{0.29}\stdpm{0.05} & \sigimprove{0.11}\stdpm{0.01} & \sigimprove{0.03}\stdpm{0.01} & 0.02\stdpm{0.01} \\
\hline
Qwen3-32B & IR & 0.26\stdpm{0.05} & 0.23\stdpm{0.05} & 0.06\stdpm{0.01} & 0.00\stdpm{0.01} & 0.00\stdpm{0.01} \\
Qwen3-32B & +IB & \sigimprove{0.41}\stdpm{0.02} & \sigimprove{0.39}\stdpm{0.02} & \sigimprove{0.11}\stdpm{0.02} & 0.02\stdpm{0.02} & 0.01\stdpm{0.01} \\
\hline
Qwen3-30B-A3B & IR & 0.40\stdpm{0.00} & 0.38\stdpm{0.00} & 0.02\stdpm{0.00} & 0.00\stdpm{0.00} & 0.00\stdpm{0.00} \\
Qwen3-30B-A3B & +IB & \sigimprove{0.53}\stdpm{0.00} & \sigimprove{0.48}\stdpm{0.00} & \sigimprove{0.12}\stdpm{0.00} & \sigimprove{0.03}\stdpm{0.00} & \sigimprove{0.02}\stdpm{0.00} \\
\hline
Qwen3-Coder-30B-A3B & IR & 0.60\stdpm{0.00} & \textbf{0.59}\stdpm{0.00} & 0.22\stdpm{0.00} & 0.05\stdpm{0.00} & 0.02\stdpm{0.00} \\
Qwen3-Coder-30B-A3B & +IB & \underline{\sigimprove{0.63}\stdpm{0.00}} & \textbf{0.59}\stdpm{0.00} & \underline{\sigimprove{0.23}\stdpm{0.00}} & 0.01\stdpm{0.00} & 0.00\stdpm{0.00} \\
\hline
DeepSeek-Coder & IR & 0.58\stdpm{0.00} & 0.50\stdpm{0.00} & 0.18\stdpm{0.00} & 0.03\stdpm{0.00} & 0.00\stdpm{0.00} \\
DeepSeek-Coder & +IB & \sigimprove{0.59}\stdpm{0.00} & \sigimprove{0.54}\stdpm{0.00} & \sigimprove{0.20}\stdpm{0.00} & \sigimprove{0.05}\stdpm{0.00} & \underline{\sigimprove{0.03}\stdpm{0.00}} \\
\hline
deepseek-v3.1 & IR & 0.28\stdpm{0.01} & 0.26\stdpm{0.02} & 0.09\stdpm{0.01} & 0.05\stdpm{0.01} & 0.02\stdpm{0.02} \\
deepseek-v3.1 & +IB & \sigimprove{0.57}\stdpm{0.08} & \sigimprove{0.55}\stdpm{0.07} & \sigimprove{0.20}\stdpm{0.03} & \underline{\sigimprove{0.09}\stdpm{0.02}} & \underline{0.03}\stdpm{0.01} \\
\hline
devstral-2512 & IR & 0.38\stdpm{0.02} & 0.33\stdpm{0.02} & 0.09\stdpm{0.01} & 0.05\stdpm{0.02} & 0.02\stdpm{0.01} \\
devstral-2512 & +IB & \sigimprove{0.69}\stdpm{0.02} & \underline{\sigimprove{0.57}\stdpm{0.01}} & \sigimprove{0.14}\stdpm{0.01} & 0.05\stdpm{0.01} & 0.02\stdpm{0.01} \\
\hline
gpt-oss-120b & IR & 0.53\stdpm{0.03} & 0.48\stdpm{0.05} & 0.21\stdpm{0.02} & 0.04\stdpm{0.01} & 0.02\stdpm{0.02} \\
gpt-oss-120b & +IB & \underline{\sigimprove{0.63}\stdpm{0.05}} & 0.56\stdpm{0.07} & \sigimprove{0.33}\stdpm{0.05} & \textbf{0.11}\stdpm{0.06} & \textbf{0.05}\stdpm{0.03} \\
\bottomrule
\end{tabular}

\end{table*}

\begin{figure*}[ht]
  % \vskip 0.2in
  \begin{center}
    \centerline{\includegraphics[width=0.75\linewidth]{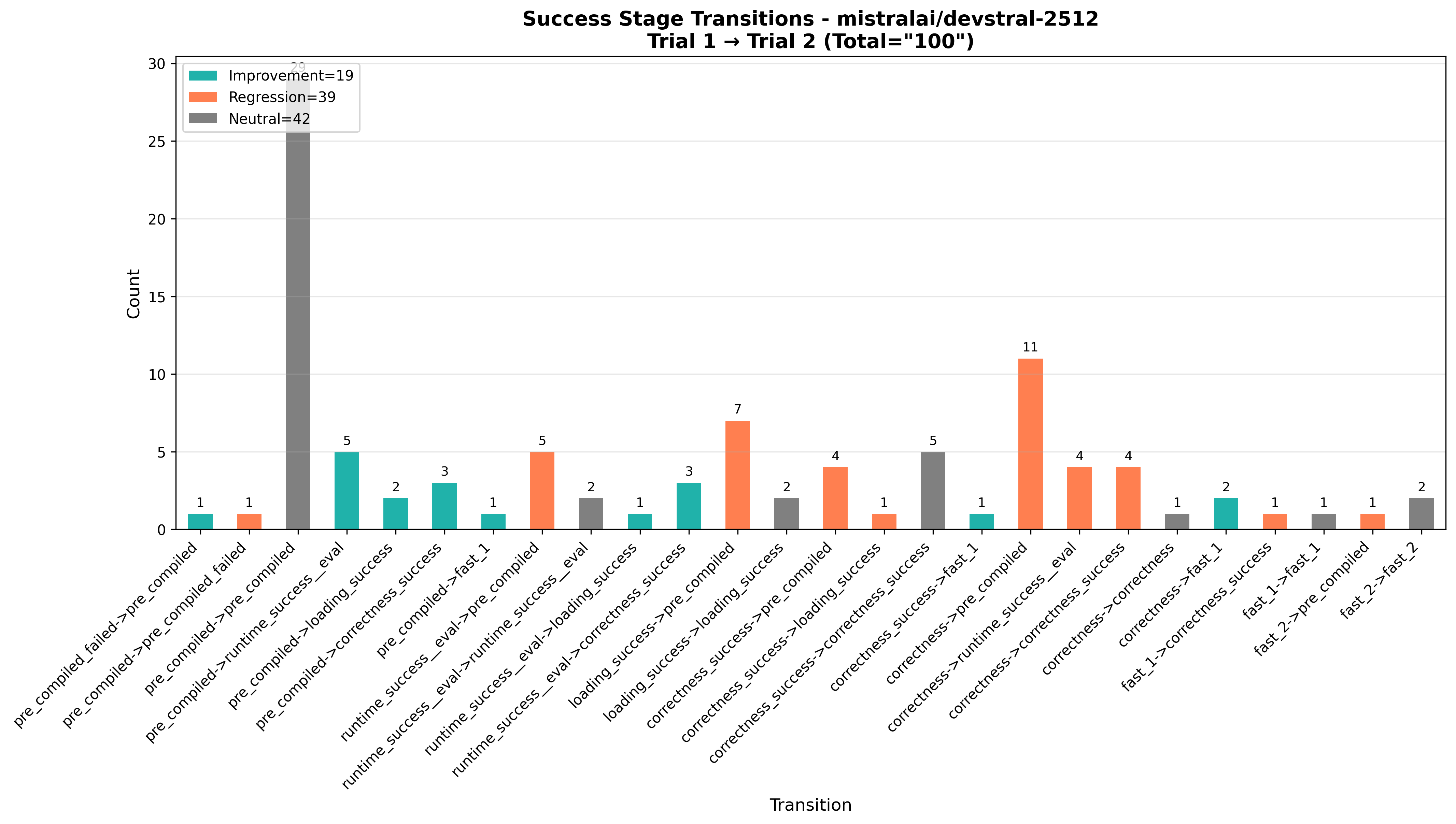}}
    \caption{
      Transition Counts for Devstral 2, from the base Trial 1 to the one based on classical iterative refinement Trial 2.
    }
    \label{fig:transitions_normal}
  \end{center}
\end{figure*}

\begin{figure*}[h]
  % \vskip 0.2in
  \begin{center}
    \centerline{\includegraphics[width=0.75\linewidth]{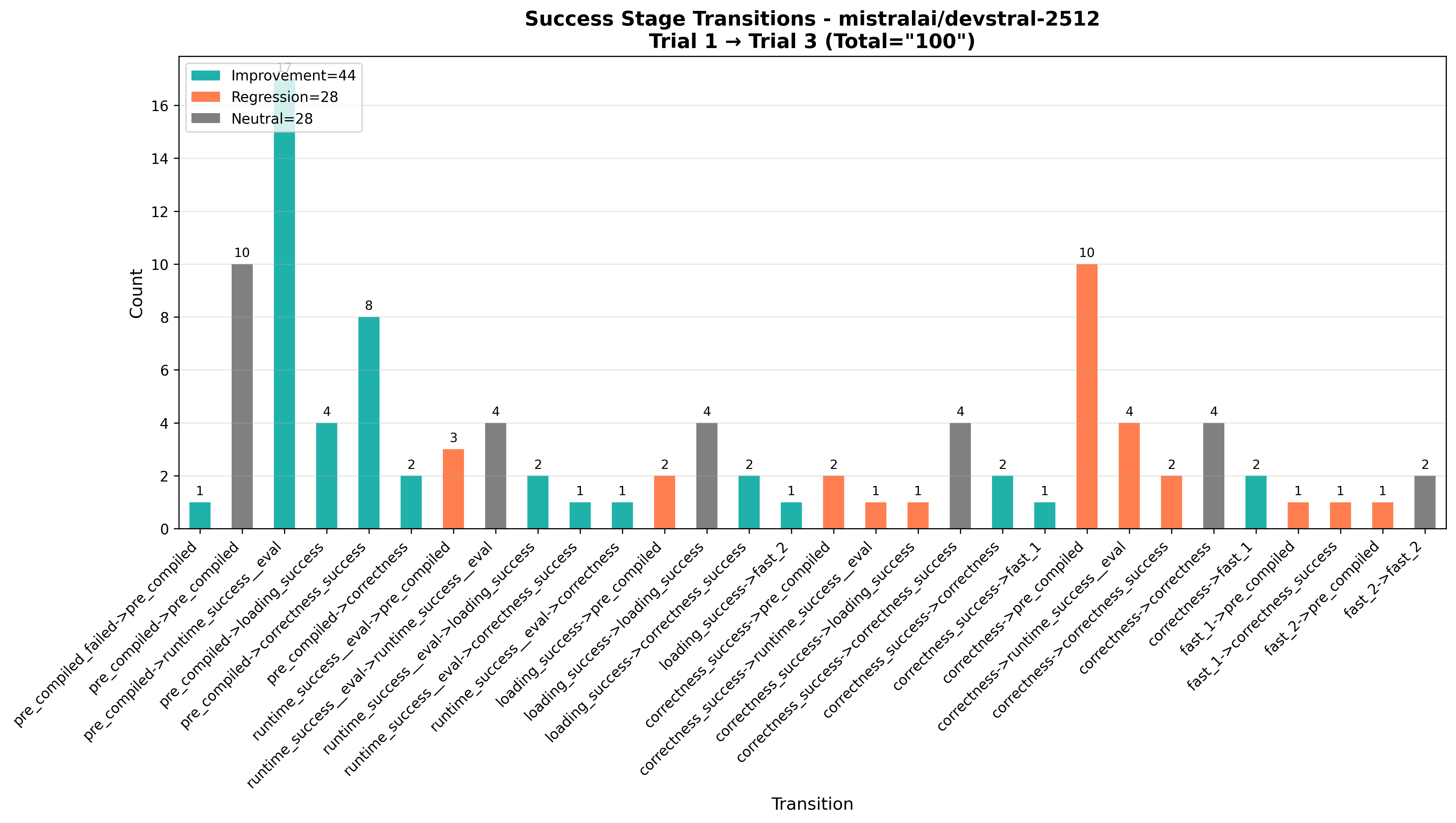}}
    \caption{
      Transition Counts for Devstral 2, from the base Trial 1 to the one based on Inductive Biases Trial 3.
    }
    \label{fig:transitions_multi}
  \end{center}
\end{figure*}

In the figure plots of the transitions Figure \ref{fig:transitions_normal} \& \ref{fig:transitions_multi}; we observe the transition counts from the previous best metric of a given sample to the new best metric. What we can see is that inductive biases have the propensity to bias the metric towards improvement; especially in the easier part of the metrics. 

\section{Impact Statement}
\label{app:impact}

This paper presents work whose goal is to advance the field of Machine Learning. There are many potential societal consequences of our work, none which we feel must be specifically highlighted here. 

The only exception to this is perhaps the fact that AI progress might be significantly limited by the implications of our hypothesis, at the same time this work would indicate a bigger role for humans in the future of AI.

\section{Additional discussion and limitations}
\label{app:discussion}
\subsection{Relation to the Feedback Information Loop hypothesis}

We hypothesized that as the feedback information loop (FIL) lengthens, purely data driven iterative refinement becomes less effective, and methods that embed inductive biases should gain an advantage. The KernelBench task has a FIL on the order of minutes, longer than the typical classification or game tasks, but far shorter than the hours or weeks envisioned for future scientific applications. Within this setting, inductive biases consistently outperforms iterative refinement, which is \textit{consistent with} the hypothesis. We also conducted a controlled experiment where adding additional domain data reduced the effect of the inductive bias, in line with the `Bitter Lesson' and further affirming that both data solves problems, but also that data needs might become inhibitive.

\subsubsection{Statistical significance for some models has low variance}
For the Qwen-A30-A3B models as well as DeepSeek-Coder, standard deviations are reported as 0.00, likely stemming from low (or non existent) variance during inference, making significance testing trivial, but also potentially less reliable for these specific models. Overall this does not change the results as we have clear statistical significance for the other results.

\section{LLM usage}
\label{app:llm_usage}
LLM usage was done for: 
\begin{enumerate}
    \item Looking material up (similar to a Web Search)
    \item Simple coding tasks (e.g. processing existing .csv files for better data analysis).
    \item Minimal rewritting help (e.g. of a paragraph or sentence).
    \item Help with Latex Table formatting.
\end{enumerate}

% \section{Technical appendices and supplementary material}
% Technical appendices with additional results, figures, graphs, and proofs may be submitted with the paper submission before the full submission deadline (see above). You can upload a ZIP file for videos or code, but do not upload a separate PDF file for the appendix. There is no page limit for the technical appendices. 

% Note: Think of the appendix as ``optional reading'' for reviewers. The paper must be able to stand alone without the appendix; for example, adding critical experiments that support the main claims to an appendix is inappropriate. 

%%%%%%%%%%%%%%%%%%%%%%%%%%%%%%%%%%%%%%%%%%%%%%%%%%%%%%%%%%%%

\newpage
\section*{NeurIPS Paper Checklist}

%%% BEGIN INSTRUCTIONS %%%
The checklist is designed to encourage best practices for responsible machine learning research, addressing issues of reproducibility, transparency, research ethics, and societal impact. Do not remove the checklist: {\bf The papers not including the checklist will be desk rejected.} The checklist should follow the references and follow the (optional) supplemental material.  The checklist does NOT count towards the page
limit. 

Please read the checklist guidelines carefully for information on how to answer these questions. For each question in the checklist:
\begin{itemize}
    \item You should answer \answerYes{}, \answerNo{}, or \answerNA{}.
    \item \answerNA{} means either that the question is Not Applicable for that particular paper or the relevant information is Not Available.
    \item Please provide a short (1--2 sentence) justification right after your answer (even for \answerNA). 
   % \item {\bf The papers not including the checklist will be desk rejected.}
\end{itemize}

{\bf The checklist answers are an integral part of your paper submission.} They are visible to the reviewers, area chairs, senior area chairs, and ethics reviewers. You will also be asked to include it (after eventual revisions) with the final version of your paper, and its final version will be published with the paper.

The reviewers of your paper will be asked to use the checklist as one of the factors in their evaluation. While \answerYes{} is generally preferable to \answerNo{}, it is perfectly acceptable to answer \answerNo{} provided a proper justification is given (e.g., error bars are not reported because it would be too computationally expensive'' or ``we were unable to find the license for the dataset we used''). In general, answering \answerNo{} or \answerNA{} is not grounds for rejection. While the questions are phrased in a binary way, we acknowledge that the true answer is often more nuanced, so please just use your best judgment and write a justification to elaborate. All supporting evidence can appear either in the main paper or the supplemental material, provided in appendix. If you answer \answerYes{} to a question, in the justification please point to the section(s) where related material for the question can be found.

IMPORTANT, please:
% \begin{itemize}
%     \item {\bf Delete this instruction block, but keep the section heading ``NeurIPS Paper Checklist"},
%     \item  {\bf Keep the checklist subsection headings, questions/answers and guidelines below.}
%     \item {\bf Do not modify the questions and only use the provided macros for your answers}.
% \end{itemize} 

%%% END INSTRUCTIONS %%%

\begin{enumerate}

\item {\bf Claims}
    \item[] Question: Do the main claims made in the abstract and introduction accurately reflect the paper's contributions and scope?
    \item[] Answer: \answerYes{} % Replace by \answerYes{}, \answerNo{}, or \answerNA{}.
    \item[] Justification: The paper clearly introduces the proof of concept of FIL hypothesis and tests it carefully, see Results section.

\item {\bf Limitations}
    \item[] Question: Does the paper discuss the limitations of the work performed by the authors?
    \item[] Answer: \answerYes{} % Replace by \answerYes{}, \answerNo{}, or \answerNA{}.
    \item[] Justification: Yes, see Appendix \ref{app:discussion}.

\item {\bf Theory assumptions and proofs}
    \item[] Question: For each theoretical result, does the paper provide the full set of assumptions and a complete (and correct) proof?
    \item[] Answer: \answerNA{} % Replace by \answerYes{}, \answerNo{}, or \answerNA{}.
    \item[] Justification: Not applicable.

    \item {\bf Experimental result reproducibility}
    \item[] Question: Does the paper fully disclose all the information needed to reproduce the main experimental results of the paper to the extent that it affects the main claims and/or conclusions of the paper (regardless of whether the code and data are provided or not)?
    \item[] Answer: \answerYes{} % Replace by \answerYes{}, \answerNo{}, or \answerNA{}.
    \item[] Justification: Methods, Experiments, and Appendix \ref{app:implementation} all explain the results. Additionally data with LLM outputs is attached (to the allowed upload limit).

\item {\bf Open access to data and code}
    \item[] Question: Does the paper provide open access to the data and code, with sufficient instructions to faithfully reproduce the main experimental results, as described in supplemental material?
    \item[] Answer: \answerYes{} % Replace by \answerYes{}, \answerNo{}, or \answerNA{}.
    \item[] Justification: Data and part of code is already attached. The full code will be released upon publication.

\item {\bf Experimental setting/details}
    \item[] Question: Does the paper specify all the training and test details (e.g., data splits, hyperparameters, how they were chosen, type of optimizer) necessary to understand the results?
    \item[] Answer: \answerYes{} % Replace by \answerYes{}, \answerNo{}, or \answerNA{}.
    \item[] Justification: Sections: Method, Experiments and Appendix \ref{app:implementation} specify these parameters.

\item {\bf Experiment statistical significance}
    \item[] Question: Does the paper report error bars suitably and correctly defined or other appropriate information about the statistical significance of the experiments?
    \item[] Answer: \answerYes{} % Replace by \answerYes{}, \answerNo{}, or \answerNA{}.
    \item[] Justification: All results tables include statistical testing using Welch-T testing and p-value of 0.05; all runs reported in the paper are N=3.

\item {\bf Experiments compute resources}
    \item[] Question: For each experiment, does the paper provide sufficient information on the computer resources (type of compute workers, memory, time of execution) needed to reproduce the experiments?
    \item[] Answer: \answerYes{} % Replace by \answerYes{}, \answerNo{}, or \answerNA{}.
    \item[] Justification: Appendix \ref{app:implementation} contains the information.
    
\item {\bf Code of ethics}
    \item[] Question: Does the research conducted in the paper conform, in every respect, with the NeurIPS Code of Ethics \url{https://neurips.cc/public/EthicsGuidelines}?
    \item[] Answer: \answerYes{} % Replace by \answerYes{}, \answerNo{}, or \answerNA{}.
    \item[] Justification: Yes, (additionally, no human study was conducted.)

\item {\bf Broader impacts}
    \item[] Question: Does the paper discuss both potential positive societal impacts and negative societal impacts of the work performed?
    \item[] Answer: \answerYes{} % Replace by \answerYes{}, \answerNo{}, or \answerNA{}.
    \item[] Justification: We don't see major impact of this work on society as of now, see Appendix \ref{app:impact} for a brief comment.
    
\item {\bf Safeguards}
    \item[] Question: Does the paper describe safeguards that have been put in place for responsible release of data or models that have a high risk for misuse (e.g., pre-trained language models, image generators, or scraped datasets)?
    \item[] Answer: \answerNA{} % Replace by \answerYes{}, \answerNo{}, or \answerNA{}.
    \item[] Justification: Not applicable.

\item {\bf Licenses for existing assets}
    \item[] Question: Are the creators or original owners of assets (e.g., code, data, models), used in the paper, properly credited and are the license and terms of use explicitly mentioned and properly respected?
    \item[] Answer: \answerYes{} % Replace by \answerYes{}, \answerNo{}, or \answerNA{}.
    \item[] Justification: See Appendix \ref{app:implementation}.

\item {\bf New assets}
    \item[] Question: Are new assets introduced in the paper well documented and is the documentation provided alongside the assets?
    \item[] Answer: \answerYes{} % Replace by \answerYes{}, \answerNo{}, or \answerNA{}.
    \item[] Justification: Software only. On release it will be documented.

\item {\bf Crowdsourcing and research with human subjects}
    \item[] Question: For crowdsourcing experiments and research with human subjects, does the paper include the full text of instructions given to participants and screenshots, if applicable, as well as details about compensation (if any)? 
    \item[] Answer: \answerNA{} % Replace by \answerYes{}, \answerNo{}, or \answerNA{}.
    \item[] Justification: No human studies were conducted.

\item {\bf Institutional review board (IRB) approvals or equivalent for research with human subjects}
    \item[] Question: Does the paper describe potential risks incurred by study participants, whether such risks were disclosed to the subjects, and whether Institutional Review Board (IRB) approvals (or an equivalent approval/review based on the requirements of your country or institution) were obtained?
    \item[] Answer: \answerNA{} % Replace by \answerYes{}, \answerNo{}, or \answerNA{}.
    \item[] Justification: No human crowdsourcing was done.

\item {\bf Declaration of LLM usage}
    \item[] Question: Does the paper describe the usage of LLMs if it is an important, original, or non-standard component of the core methods in this research? Note that if the LLM is used only for writing, editing, or formatting purposes and does \emph{not} impact the core methodology, scientific rigor, or originality of the research, declaration is not required.
    %this research? 
    \item[] Answer: \answerYes{} % Replace by \answerYes{}, \answerNo{}, or \answerNA{}.
    \item[] Justification: Both in OpenReview and in the paper, see Appendix \ref{app:llm_usage}.

\end{enumerate}

\end{document}